\documentstyle[11pt]{article}
\begin{document}
\title{{\bf Decentralized Failure Diagnosis of Stochastic Discrete Event Systems }
\thanks{This work was supported in part by the National
Natural Science Foundation under Grant 90303024 and Grant
60573006, the Higher School Doctoral Subject Foundation of
Ministry of Education under Grant 20050558015, and the Guangdong
Province Natural Science Foundation under Grant 020146 and Grant
031541 of China.}}
\author{Fuchun Liu$^{1,2}$, Daowen Qiu$^{1}$, Hongyan Xing$^{1,2}$, and Zhujun Fan$^{3}$\\
{\footnotesize   $^{1}$Department of Computer Science, Zhongshan
University,
Guangzhou 510275, China}\\
{\footnotesize  $^{2}$Faculty of Applied Mathematics, Guangdong
University of Technology, Guangzhou 510090, China}\\
{\footnotesize   $^{3}$Department of Mathematics, Zhongshan
University, Guangzhou 510275, China} }

\date{  }
\maketitle

\begin{center}
\begin{minipage}{150mm}

{\bf Abstract:}  Recently, the diagnosability of {\it stochastic
discrete event systems} (SDESs) was  investigated in the
literature, and, the failure diagnosis considered was {\it
centralized}. In this paper, we propose an approach to {\it
decentralized} failure diagnosis of SDESs, where the stochastic
system uses multiple local diagnosers to detect failures and each
local diagnoser possesses its own information. In a way, the
centralized failure diagnosis of SDESs can be viewed as a special
case of the decentralized failure diagnosis presented in this
paper with only one projection. The main contributions are as
follows: (1) We formalize the notion of codiagnosability for
stochastic automata, which means that a failure can be detected by
at least one local stochastic diagnoser within a finite delay. (2)
We construct a codiagnoser from a given stochastic automaton with
multiple projections, and the codiagnoser associated with the
local diagnosers is used to test codiagnosability condition of
SDESs.  (3) We deal with a number of basic properties of the
codiagnoser. In particular, a necessary and sufficient condition
for the codiagnosability of SDESs is presented. (4) We give a
computing method in detail to check whether codiagnosability is
violated.  And (5) some examples are described to illustrate the
applications of the codiagnosability and its computing method.

\par
\hskip 5mm


\end{minipage}
\end{center}
\smallskip

\section*{I. Introduction}

A {\it discrete event system} (DES) is a dynamical system whose
state space is discrete and whose states can only change when
certain events occur [1, 2], which has been successfully applied
to provide a formal treatment of many technological and
engineering systems [1]. In order to guarantee performance to a
reliable system, the control engineers should design a system that
runs safely within its normal boundaries. Therefore, failure
diagnoses in DESs, which are to detect and isolate the
unobservable fault events occurring in a system within a finite
delay, are of practical and theoretical importance, and have
received considerable attention in recent years [1, 3-34].

In the past a long time, most of the research works on failure
diagnosis of DESs in the literature focused on centralized failure
diagnosis usually [3, 5, 9-17, 19, 20, 23-27, 29, 31-34]. Many
large complex systems, however, are physically distributed systems
in nature [6, 7, 21], where information diagnosed is decentralized
and there are several local sites, in which sensors report their
data and diagnosers run at each site processing the local
observation. Therefore, in recent years, more and more research
works have devoted to decentralized failure diagnosis [4, 6-8, 18,
21, 22, 28, 30].

As we know, the classical DES models cannot distinguish between
strings or states that are highly probable and those that are less
probable, and the notion that a failure can be diagnosed after a
finite delay is ``all-or-nothing" [31]. Stochastic automata, as a
natural generalization for deterministic automata of different
types, are a more precise formulation of the general DES models,
in which a probabilistic structure is appended to estimate the
likelihood of specific events occurring [31]. An introduction to
the theory of stochastic automata can be found in [2].

More recently, by generalizing the diagnosability of classical
DESs [25, 26] to the setting of {\it stochastic discrete event
systems} (SDESs), the diagnosability of SDESs was interestingly
dealt with by J. Lunze and J. Schr\"{o}der [16], D. Thorsley and
D. Teneketzis [31]. In [16], the diagnostic problem was
transformed into an observation problem, and the diagnosability
was obtained by an extension of an observation algorithm. In [31],
the notions of A- and AA-diagnosability for stochastic automata
were defined, which were weaker than those for classical automata
introduced by Sampath {\it et al} [25, 26], and they presented a
necessary and sufficient condition for the diagnosability of
SDESs. However,  the failure diagnosis considered in SDESs was
{\it centralized}. Therefore, motivated by the importance of
decentralized failure diagnosis, our goal is to deal with the
decentralized failure diagnosis for SDESs.

In this paper, we formalize the approach to decentralized failure
diagnosis in SDESs by introducing the notion of codiagnosability.
The centralized failure diagnosis in SDESs [16, 31] can be viewed
as a special case of the decentralized failure diagnosis presented
in this paper with only one projection. Roughly speaking, a
language generated by a stochastic automaton is said to be
codiagnosable under some local projections if, after a failure
event occurs, there exists at least one local site such that the
probability of non-diagnosing failure is sufficiently small within
a finite delay. By constructing a codiagnoser from a given
stochastic automaton with multiple projections, we can use the
codiagnoser associated with the local diagnosers to test
codiagnosability condition of SDESs.  As well, a number of basic
properties of the codiagnoser is investigated. In particular, a
necessary and sufficient condition for the codiagnosability of
SDESs is presented, which generalizes the result of classical DESs
dealt with by W. Qiu and R. Kumar [21]. Furthermore, we propose a
computing method in detail to check whether codiagnosability is
violated. Finally, some examples are described to illustrate the
applications of the codiagnosability and its computing method.

This paper is organized as follows. Section II serves to recall
some related concepts and notations concerning failure diagnosis
of DESs and SDESs. In Section III, we introduce a definition of
the codiagnosability of SDESs. The codiagnoser used to detect
failure in SDESs is constructed. In Section IV, some main
properties of codiagnoser are investigated. In particular, a
necessary and sufficient condition for the codiagnosability of
SDESs is presented. As well, we give a computing method in detail
to check whether codiagnosability is violated, according to the
codiagnoser and the local stochastic diagnosers. Furthermore, some
examples are provided to illustrate the condition of the
codiagnosability for SDESs.  Finally, in Section V, we summarize
the main results of the paper and address some related issues.

\section*{II. Notations and preliminaries}
In this section, we present some preliminaries concerning
stochastic automata and centralized failure diagnosis of SDESs.
For more details on SDESs, we can refer to [2,16,31].

\subsection*{{\it A. Stochastic Automata }}
A stochastic automaton is a finite state machine (FSM) with a
probabilistic structure.

{\it Definition 1[31]:}  {\it A stochastic automaton} is a type of
systems with a quadruple
\begin{equation}
G=(Q,\Sigma,\eta, q_{0}),
\end{equation}
where $Q$ is a finite state space; $q_{0}\in Q $ is the initial
state; $\Sigma$ is a finite set of events; $\eta:
Q\times\Sigma\times Q\rightarrow [0,1] $ is a state transition
function: for any $q, q^{'}\in Q $ and any $\sigma\in\Sigma$,
$\eta(q, \sigma, q^{'})$ represents the probability that a certain
event $\sigma$ will occur, together with transferring the state of
the machine from a given state $q$ to the specified state
$q^{'}$. For example, $\eta(q, \sigma, q^{'})=0.7$ means that, if
the machine is in state $q$, then with probability 0.7 event
$\sigma$ will occur, together with transferring to state $q^{'}$.

For the sake of simplicity, we assume like [31] that, for a given
state $q\in Q $ and a given event $\sigma\in\Sigma$, there exists
at most one state $q^{'}\in Q $ such that $\eta(q, \sigma,
q^{'})>0$. Therefore, we can sometimes use the symbol $\eta(q,
\sigma)$ instead of $\eta(q, \sigma, q^{'})$ hereafter. Moreover,
we recursively define
\begin{equation}
Pr(\sigma\mid q)=\eta(q, \sigma), \hskip 7mm Pr(s\sigma\mid q)
=Pr(s\mid q)\eta(q^{'}, \sigma),\end{equation}
 where $\sigma\in\Sigma$,
$s\in\Sigma^{*}$ and $\eta(q,s,q^{'})>0$. Intuitively,
$Pr(\sigma\mid q)$ or $Pr(s\sigma\mid q)$ represents the
probability of event $\sigma$ or string $s\sigma$ being the next
event or string when the system is in state $q$. We can simply
denote them by $Pr(\sigma)$ and $Pr(s\sigma)$, respectively, if no
confusion results.

Some events in $\Sigma$ occurring can be observed by the sensors,
while the rest are unobservable. That is,
$\Sigma=\Sigma_{o}\cup\Sigma_{uo}$, where $\Sigma_{o}$ represents
the set of observable events and $\Sigma_{uo}$ the set of
unobservable events. Let $\Sigma_{f}\subseteq \Sigma$ denote the
set of failure events which are to be diagnosed. Without loss of
generality, we can assume that $\Sigma_{f}\subseteq \Sigma_{uo}$,
as [24, 28, 29, 34]. And $\Sigma_{f}$ is partitioned into
different failure types
$$\Sigma_{f}=\Sigma_{f_{1}}\cup\Sigma_{f_{2}}\cup\ldots\cup\Sigma_{f_{m}}.
$$
If a failure event $\sigma\in\Sigma_{f_{i}}$ occurs, we will say
that a failure type $F_{i}$ has occurred.

The language generated by a stochastic automaton $G$, denoted by $
L(G)$, or $L$ for simplicity, is the set of all finite strings
with positive probability. That is,
\begin{equation}
 L=\left\{s\in \Sigma^{*}: (\exists q\in Q)\eta(q_{0}, s, q)>0\right\}.
\end{equation}
A trace $s\in L$ is called to be a deadlocking trace if no further
continuations exist after it is in $L$, i.e.,
$\left\{s\right\}\Sigma^{*}\cap L=\left\{s\right\}$. Without loss
of generality, we assume that $L$ is deadlock-free. Otherwise, we
can extend each deadlocking trace by an unbounded sequence of a
newly added event that is unobservable to all diagnosers. This
will make the language deadlock-free without altering any
properties of diagnosability [12, 21].

When a system execution is observed by an observation, events
executed by the system are filtered and the unobservable events
are erased by a projection.

{\it Definition 2:}  A {\it projection} $P:\Sigma^{*}\rightarrow
\Sigma_{o}^{*}$ is defined: $P(\epsilon)=\epsilon$, and for any $
\sigma\in \Sigma $,\hskip 2mm $ s\in \Sigma^{*} $, $
P(s\sigma)=P(s)P(\sigma)$, where
\begin{equation}
\begin{array}{lll}
P(\sigma)=\left\{
\begin{array}{ll}
\sigma,    \hskip 3mm&{\rm if}  \hskip 3mm \sigma\in \Sigma_{o},\\
\epsilon, \hskip 3mm&{\rm if}  \hskip 3mm \sigma\in \Sigma_{uo}.
\end{array}
\right.
\end{array}
\end{equation}
And the inverse projection of string $y\in \Sigma^{*}$ is defined
as
\begin{equation}
P^{-1}(y)=\left\{s \in L: P(s)=y \right\}.
\end{equation}

We further need some notations. For string $s\in \Sigma^{*}$,
$\overline{s}$ and $s_{f}$ denote the prefix-closure of $s$ and
the final event of string $s$, respectively. We define
\begin{equation}
 L/s = \left\{t \in
\Sigma^{*}: st\in L\right\},
\end{equation}
\begin{equation}
\Psi(\Sigma_{f_{i}})= \left\{s\sigma_{f} \in L: \sigma_{f}\in
\Sigma_{f_{i}}\right\},
\end{equation}
\begin{equation}
 L(G,q)=\left\{s\in
\Sigma^{*}: (\exists q^{'}\in Q)\eta(q, s, q^{'})>0\right\}.
\end{equation}
Intuitively, $ L/s $ represents the set of continuations of the
string $s$, $ \Psi(\Sigma_{f_{i}})$ denotes the set of all traces
of $L$ that end in a failure event belonging to the class
$\Sigma_{f_{i}}$, and $L(G,q)$ denotes the set of all traces that
originate from state $q\in Q$.

\subsection*{{\it B. Centralized Failure Diagnosis of SDESs}}
Before discussing the decentralized failure diagnosis of SDESs, we
recall the centralized failure diagnoses of SDESs investigated in
[31].

{\it Definition 3 [31]:} Let $L$ be a language generated by a
stochastic automaton $G=(Q,\Sigma,\eta, q_{0})$ and let
$P:\Sigma^{*}\rightarrow \Sigma_{o}^{*}$ be a projection.  $L$ is
said to be {\it $A$-diagnosable with respect to $P$} if
\begin{equation}
\begin{array}{ll}
(\forall \epsilon>0)(\exists n_{0}\in {\bf N})(\forall s\in
\Psi(\Sigma_{f_{i}})\wedge n\geq
n_{0})\\
\{Pr(t:D(st)=0 \mid t\in L/s \wedge
\parallel t
\parallel=n)<\epsilon\}
\end{array}
\end{equation}
where the diagnosability condition function $D:
\Sigma^{*}\rightarrow \left\{0,1\right\}$ is defined as follows:
\begin{equation}
D(st)=\left\{
\begin{array}{ll}
1,    \hskip 2mm&{\rm if}  \hskip 2mm \omega\in P^{-1}[P(st)]\Rightarrow\Sigma_{f_{i}}\in \omega,\\
0, \hskip 2mm&{\rm otherwise}.
\end{array}
\right.
\end{equation}
Roughly speaking, $L$ being $A$-diagnosable means that after a
failure occurs, the probability that a string  cannot be detected
is sufficiently small within a finite delay. For simplicity, we
will call it diagnosable instead of $A$-diagnosable.

The stochastic diagnoser $G_{d}$ constructed in [31] is as
follows:
\begin{equation}
G_{d}=(Q_{d}, \Sigma_{o}, \delta_{d}, \chi_{0}, \Phi, \phi_{0}),
\end{equation}
where $Q_{d}$ is the set of states of the diagnoser with initial
state $\chi_{0}=\left\{(q_{0}, N)\right\}$, $\Sigma_{o}$ is the
set of observable events, $\delta_{d}$ is the transition function
of the diagnoser, $\Phi$ is the set of probability transition
matrices, and $\phi_{0}$ is the initial probability mass function
on $\chi_{0}$.

Some concepts concerning finite state Markov chain are used to
derive the results of [31], and we briefly recall them. Suppose
that $x$ and $y$ are two states of a Markov chain. The symbol
$\rho_{xy}$ represents the probability that if the Markov chain is
in state $x$, it will visit state $y$ at some point in the future.
For a state $x$, if $\rho_{xx}=1$, then $x$ is called  a {\it
recurrent state}. Otherwise, if $\rho_{xx}<1$, then $x$ is called
 a {\it transient state}.

We now quote the basic properties related to transient or
recurrent states from [31].

{\it Lemma 1 (Lemma 1 in [31]):} Let $\Gamma$ be the set of
transient states of a Markov chain and let $x$ be an arbitrary
state of the chain. Then for any $t\in L(G,x)$ and any
$\epsilon>0$, there exists $n\in {\bf N}$ such that
\begin{equation}
Pr(t: \delta(x, t)\in \Gamma\mid t\in L(G,x) \wedge
\parallel t
\parallel=n)<\epsilon.
\end{equation}

{\it Lemma 2 (Property 4 in [31]):} All components reachable from
a recurrent state bearing the label $F_{i}$ in an
$F_{i}$-uncertain element of $Q_{d}$ are contained in
$F_{i}$-uncertain elements.

Using the stochastic diagnoser $G_{d}$, Thorsley and Teneketzis
[31] presented a necessary and sufficient condition for
diagnosability as follows:

{\it Lemma 3 (Theorem 3 in [31]):}  A language $L$ generated by a
stochastic automaton $G$ is diagnosable, if and only if, every
logical element of its diagnoser $G_{d}$ containing a recurrent
component bearing the label $F$ is $F$-certain.

\section*{III.  Codiagnosability and  Codiagnoser for SDESs}
In order to illustrate the solution to the decentralized failure
diagnosis problem, we make the following assumptions about the
stochastic automaton $G=(Q,\Sigma,\eta, q_{0})$ as [31]:

({\it A1}): The language $L=L(G)$ is live. That is to say, for any
$q\in Q$,
\begin{equation}
\sum_{q^{'}\in Q} \sum_{\sigma\in \Sigma} \eta(q, \sigma,
q^{'})=1.
\end{equation}

({\it A2}): There does not exist any cycle of unobservable events,
i.e., $$(\exists n_{0}\in {\bf N})(\forall ust\in L)[( s\in
\Sigma_{uo}^{*})\Rightarrow (\parallel s \parallel\leq n_{0})].$$
Intuitively, assumption ({\it A1}) means that the sum of the
probability of all transitions from each state is equal to one,
which indicates that transitions will continue to occur in any
state. Assumption ({\it A2}) ensures that $G$ does not generate
arbitrarily long sequences of unobservable events, because failure
diagnosis is based on observable transitions of the system.

\subsection*{{\it A. Approaches to Defining Codiagnosability for SDESs}}
In this subsection, we consider decentralized failure diagnosis
where there are $m$ local diagnosers to detect system $G$. The $m$
local diagnosers are assumed to be independent, namely, without
communicating their observations each other [15, 24-26, 31]. From
the $m$ local projections $P_{i}:\Sigma^{*}\rightarrow
\Sigma_{o,i}^{*}$ defined as Definition 2, where $i=1,
2,\cdots,m$, we can obtain the global projection
$P:\Sigma^{*}\rightarrow \Sigma_{o}^{*}$,  in which
\begin{equation}
\Sigma_{o}=\Sigma_{o,1}\cup\Sigma_{o,2}\cup\cdots\Sigma_{o,m}.
\end{equation}

Now let us give the definition of codiagnosability for SDESs.

{\it Definition 4:} Let $G=(Q,\Sigma,\eta, q_{0})$ be a stochastic
automaton, $L=L(G)$. Assume there are $m$ local projections
$P_{i}:\Sigma^{*}\rightarrow \Sigma_{o,i}^{*}$, where $i=1,
2,\cdots,m$.  Then $L$ is said to be codiagnosable with respect to
$\left\{P_{i}\right\}$ if
\begin{equation}
\begin{array}{ll}
(\forall \epsilon>0)(\exists n_{i}\in {\bf N})(\forall s\in
\Psi(\Sigma_{f_{i}})\wedge n\geq n_{i}) (\exists j\in \left\{1,
2,\cdots,m\right\})\\ \{Pr(t:D_{j}(st)=0 \mid t\in L/s \wedge
\parallel t
\parallel=n)<\epsilon\}
\end{array}
\end{equation}
where for each $j\in \left\{1, 2,\cdots,m\right\}$, the
diagnosability condition function $D_{j}: \Sigma^{*}\rightarrow
\left\{0,1\right\}$ is defined by Definition 3, i.e.,
\begin{equation}
D_{j}(st)=\left\{
\begin{array}{ll}
1,    \hskip 2mm&{\rm if}
\hskip 2mm \omega\in P_{j}^{-1}[P_{j}(st)]\Rightarrow\Sigma_{f_{i}}\in \omega,\\
0, \hskip 2mm&{\rm otherwise}.
\end{array}
\right.
\end{equation}

Intuitively, $L$ being codiagnosable means that, for any a trace
$s$ that ends in a failure event belonging to $\Sigma_{f_{i}}$ and
for any a sufficiently long continuation $t$ of $s$, there exists
at least one site $j$ such that, the probability that the $jth$
diagnoser cannot detect the failure among the traces
indistinguishable from $st$ for site $j$ is sufficiently small
within a finite delay.

{\it Remark 1:} Comparing with Definition 3, we know that
diagnosability of the centralized system [31] can be viewed as a
special case of the codiagnosability of the decentralized system
with $m=1$.

 {\it Example 1.} Consider the stochastic
automaton $G=(Q,\Sigma,\eta, q_{0})$ described by Fig.1, where
$Q=\{q_{0}, q_{1},\ldots, q_{6}\}$, $q_{0}$ is the initial state,
$\Sigma=\{a, b, c, d, \sigma_{uo}, \sigma_{f}\}$, and the set of
failure events $\Sigma_{f}=\{\sigma_{f}\}$. Assume that there are
two local projections $P_{i}:\Sigma^{*}\rightarrow
\Sigma_{o,i}^{*}$, where $\Sigma_{o, 1}=\{a, b\}$, $\Sigma_{o,
2}=\{a, c\}$, $i=1, 2$.

\setlength{\unitlength}{0.1cm}
\begin{picture}(150,60)

\put(15,35){\circle{5}\makebox(-10,0){$0$}}
\put(45,35){\circle{5}\makebox(-10,0){$4$}}
\put(45,50){\circle{5}\makebox(-10,0){$1$}}
\put(75,50){\circle{5}\makebox(-10,0){$2$}}
\put(105,50){\circle{5}\makebox(-10,0){$3$}}
\put(45,20){\circle{5}\makebox(-10,0){$5$}}
\put(75,20){\circle{5}\makebox(-10,0){$6$}}

\put(5,35){\vector(1,0){7}} \put(17,37){\vector(2,1){25.3}}
\put(17,33){\vector(2,-1){25.3}} \put(17.6,35){\vector(1,0){25}}
\put(47.6,50){\vector(1,0){25}} \put(77.6,50){\vector(1,0){25}}
\put(47.6,20){\vector(1,0){25}}

\put(75,58){\circle{11}} \put(80.5,58){\vector(0,-1){1}}
\put(45,12){\circle{11}} \put(50.5,12){\vector(0,1){1}}
\put(53,35){\circle{11}} \put(58.7,35){\vector(0,-1){1}}
\put(113,50){\circle{11}} \put(118.7,50){\vector(0,-1){1}}
\put(83,20){\circle{11}} \put(88.7,20){\vector(0,-1){1}}

\put(25,45){\makebox(0,0)[c]{$(d, 0.5)$}}
\put(58,53){\makebox(0,0)[c]{$(\sigma_{f}, 1)$}}
\put(87,59){\makebox(0,0)[c]{$(a, 0.7)$}}
\put(90,53){\makebox(0,0)[c]{$(c, 0.3)$}}
\put(125,50){\makebox(0,0)[c]{$(a, 1)$}}
\put(34,38){\makebox(0,0)[c]{$(\sigma_{uo}, 0.2)$}}
\put(64,35){\makebox(0,0)[c]{$(a, 1)$}}

\put(25,24){\makebox(0,0)[c]{$(\sigma_{f}, 0.3)$}}
\put(60,23){\makebox(0,0)[c]{$(b, 0.2)$}}
\put(95,20){\makebox(0,0)[c]{$(a, 1)$}}
\put(58,12){\makebox(0,0)[c]{$(a, 0.8)$}}
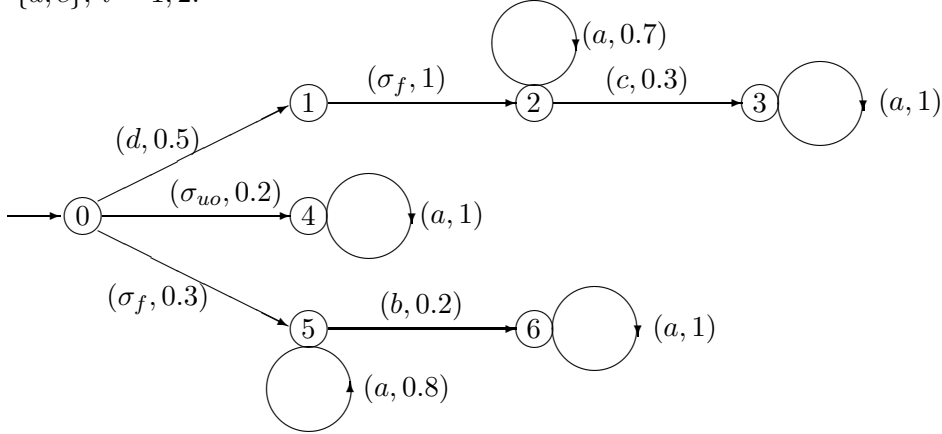
\put(45,0){\makebox(45,0)[c]{{\footnotesize Fig. 1. Stochastic
automaton of Example 1. }}}

\end{picture}

We assert that the language $L=L(G)$ is codiagnosable. In fact,
for any $ s\in \Psi(\Sigma_{f})$, i.e., $s=\sigma_{f}$ or
$s=d\sigma_{f}$, we verify its codiagnosability as follows.

{\it Case 1.} If $s=\sigma_{f}$, then for any $t\in L/s$ and
$\parallel t \parallel=n$, either $t=a^{n-k-1}ba^{k}$ (where
$0\leq k\leq n-1$) or $t=a^{n}$, and we can take the first
diagnoser to detect the failure.

When $t=a^{n-k-1}ba^{k}$, we have $$
P_{1}^{-1}[P_{1}(st)]=\left\{\sigma_{f}a^{n-k-1}ba^{k}: 0\leq
k\leq n-1 \right\}.$$ Due to $\sigma_{f}\in
\sigma_{f}a^{n-k-1}ba^{k}$ for all $k\in [0, n-1]$, we get
$D_{1}(st)=1$. Therefore, the failure is diagnosed. When
$t=a^{n}$, we have $$
P_{1}^{-1}[P_{1}(st)]=\left\{d\sigma_{f}a^{n}, \sigma_{f}a^{n},
\sigma_{uo}a^{n}, d\sigma_{f}a^{n-k}ca^{k}: 0\leq k\leq n-1
\right\}.$$ Due to $\sigma_{f}\notin \sigma_{uo}a^{n}$, we obtain
$D_{1}(st)=0$. In this case, $Pr(t)=Pr(a^{n})=0.8^{n}$,  and with
$n$ increasing, the probability that is not diagnosable approaches
to zero.

{\it Case 2.} If $s=d\sigma_{f}$, then for any $t\in L/s$ and
$\parallel t \parallel=n$, either $t=a^{n-k-1}ca^{k}$ (where
$0\leq k\leq n-1$) or $t=a^{n}$, and we can take the second
diagnoser to detect the failure.

When $t=a^{n-k-1}ca^{k}$, we have $$
P_{2}^{-1}[P_{2}(st)]=\left\{d\sigma_{f}a^{n-k-1}ca^{k}: 0\leq
k\leq n-1 \right\}.$$  Due to $\sigma_{f}\in
d\sigma_{f}a^{n-k-1}ca^{k}$ for all $k\in [0, n-1]$, we get
$D_{2}(st)=1$. Therefore, the failure is diagnosed. When
$t=a^{n}$, we have $$
P_{2}^{-1}[P_{2}(st)]=\left\{d\sigma_{f}a^{n}, \sigma_{f}a^{n},
\sigma_{uo}a^{n}, \sigma_{f}a^{n-k}ba^{k}: 0\leq k\leq n-1
\right\}.$$ Due to $\sigma_{f}\notin \sigma_{uo}a^{n}$, we obtain
$D_{2}(st)=0$. In this case, $Pr(t)=Pr(a^{n})=0.7^{n}$, and with
$n$ increasing, the probability that is not diagnosable approaches
to zero.

By Definition 4, Case 1 and Case 2 indicate $L$ is codiagnosable.
\hfill $\Box$

Before constructing the codiagnoser for SDESs, we first give two
propositions for the condition of non-codiagnosability, which can
be straight obtained from Definition 4.

{\it Proposition 1:} Let $G=(Q,\Sigma,\eta, q_{0})$ be a
stochastic automaton, $L=L(G)$. Assume there are $m$ local
projections $P_{i}:\Sigma^{*}\rightarrow \Sigma_{o,i}^{*}$, where
$i=1, 2,\cdots,m$. If there exist $i_{0}\in \left\{1,
2,\cdots,m\right\}$, such that $L$ is diagnosable with respect to
$P_{i_{0}}$, then $L$ is codiagnosable with respect to
$\left\{P_{i}\right\}$.

{\it Proof:} If there exists $i_{0}\in \left\{1,
2,\cdots,m\right\}$, such that $L$ is diagnosable with respect to
$P_{i_{0}}$, then from Definition 3,  the following holds:
\begin{equation}
\begin{array}{ll}
(\forall \epsilon>0)(\exists n_{0}\in {\bf N})(\forall s\in
\Psi(\Sigma_{f_{i}})\wedge n\geq
n_{0})\\
\{ Pr(t:D_{i_{0}}(st)=0 \mid t\in L/s \wedge
\parallel t
\parallel=n)<\epsilon\}.
\end{array}
\end{equation}
Therefore, we have
\begin{equation}
\begin{array}{ll}
(\forall \epsilon>0)(\exists n_{i}=n_{0})(\forall s\in
\Psi(\Sigma_{f_{i}})\wedge n\geq n_{i})(\exists j=i_{0})\\
\{Pr(t:D_{j}(st)=0 \mid t\in L/s \wedge
\parallel t
\parallel=n)<\epsilon\}.
\end{array}
\end{equation}
It indicates that $L$ is codiagnosable with respect to
$\left\{P_{i}: i=1, 2,\cdots,m\right\}$ by Definition 4.\hfill
$\Box$

{\it Remark 2:} This proposition shows that a system can detect
all of the failure strings if a local diagnoser can detect them.
However, the inverse proposition does not always hold. That is,
there exists the case that a system can still detect all of the
failure strings even if all of the local diagnosers cannot detect
the failures. Example 2 verifies this view.

{\it Example 2.} For the stochastic automaton $G=(Q,\Sigma,\eta,
q_{0})$ as in Example 1, we have known that the language $L$ is
codiagnosable from Example 1. However, in the following we prove
that $L$ is not diagnosable with respect to $P_{1}:
\Sigma^{*}\rightarrow \Sigma_{o,1}^{*}$, neither is $L$
diagnosable with respect to $P_{2}: \Sigma^{*}\rightarrow
\Sigma_{o,2}^{*}$, where $\Sigma_{o, 1}=\{a, b\}$ and $\Sigma_{o,
2}=\{a, c\}$.

In fact, for the first projection $P_{1}: \Sigma^{*}\rightarrow
\Sigma_{o,1}^{*}$, we can take $\epsilon=0.2$, $s=d\sigma_{f}\in
\Psi(\Sigma_{f})$ and $t=aca^{n-2}\in L/s $, then,
$$ P_{1}^{-1}[P_{1}(st)]=\left\{d\sigma_{f}a^{n-1},
\sigma_{f}a^{n-1}, \sigma_{uo}a^{n-1}, d\sigma_{f}a^{n-1-k}ca^{k}:
0\leq k\leq n-1  \right\}.$$ Because $\sigma_{f}\notin
\sigma_{uo}a^{n-1}$, the diagnosability condition function
$D_{1}(st)=0$. But
\begin{equation}
Pr(t:D_{1}(st)=0 \mid t\in L/s \wedge
\parallel t
\parallel=n)=Pr(aca^{n-2})=0.21>\epsilon.
\end{equation}

Similarly, for the second projection $P_{2}: \Sigma^{*}\rightarrow
\Sigma_{o,2}^{*}$, we can take $\epsilon=0.1$, $s=\sigma_{f}\in
\Psi(\Sigma_{f})$ and $t=aba^{n-2}\in L/s $, then,
$$ P_{2}^{-1}[P_{2}(st)]=\left\{d\sigma_{f}a^{n-1},
\sigma_{f}a^{n-1}, \sigma_{uo}a^{n-1}, \sigma_{f}a^{n-1-k}ba^{k}:
0\leq k\leq n-1  \right\}.$$ Because $\sigma_{f}\notin
\sigma_{uo}a^{n-1}$, the diagnosability condition function
$D_{2}(st)=0$. But
\begin{equation}
Pr(t:D_{2}(st)=0 \mid t\in L/s \wedge
\parallel t
\parallel=n)=Pr(aba^{n-2})=0.16>\epsilon.
\end{equation}

Eqs. (19, 20) indicate that  $L$ is not diagnosable with respect
to $P_{1}$, neither is $L$ diagnosable with respect to $P_{2}$.
\hfill $\Box$

{\it Proposition 2:} Let $G=(Q,\Sigma,\eta, q_{0})$ be a
stochastic automaton, $L=L(G)$. Assume there are $m$ local
projections $P_{i}:\Sigma^{*}\rightarrow \Sigma_{o,i}^{*}$, where
$i=1, 2,\cdots,m$. $L$ is not codiagnosable with respect to
$\left\{P_{i}\right\}$, if and only if,
\begin{equation}
\begin{array}{ll}
(\exists \epsilon>0)(\forall n_{i}\in {\bf N})(\exists s\in
\Psi(\Sigma_{f_{i}})\wedge n\geq n_{i})(\exists t\in L/s)
(\forall j\in \left\{1, 2,\cdots,m\right\})\\
\{Pr(t:D_{j}(st)=0 \mid t\in L/s \wedge
\parallel t
\parallel=n)\geq\epsilon\}.
\end{array}
\end{equation}

{\it Proof:} It can be readily obtained from Definition 4. \hfill
$\Box$

{\it Remark 3:} If $m=2$ and
$\Sigma_{f}=\left\{\sigma_{f}\right\}$, then $L$ being not
codiagnosable means the following: there exists $\epsilon>0$, such
that for any $ n_{i}\in {\bf N}$, there exist $s\in
\Psi(\Sigma_{f})$, $t\in L/s $,  $\omega_{1}\in
P_{1}^{-1}[P_{1}(st)]$, and $\omega_{2}\in P_{2}^{-1}[P_{2}(st)]$,
satisfying  $\sigma_{f}\notin \omega_{1}$, $\sigma_{f}\notin
\omega_{2}$, and
\begin{equation}
Pr(t:D_{1}(st)=0 \mid t\in L/s \wedge
\parallel t
\parallel=n)\geq\epsilon,
\end{equation}
\begin{equation}
Pr(t:D_{2}(st)=0 \mid t\in L/s \wedge
\parallel t
\parallel=n)\geq\epsilon.
\end{equation}

\subsection*{{\it B. Construction of Codiagnoser from a Stochastic Automaton}}
Firstly, we will define a logical finite state automaton from a
given stochastic automaton.

{\it Definition 5:} Let $G=(Q,\Sigma,\eta, q_{0})$ be a given
stochastic automaton. The {\it deterministic finite automaton
(DFA) deduced by $G$ } is defined as $G^{'}=(Q, \Sigma, \delta,
q_{0})$ with the same sets of states and events, but the partial
transition function $\delta: Q\times\Sigma \rightarrow Q$ in
$G^{'}$ is determined by probability function $\eta:
Q\times\Sigma\times Q\rightarrow [0,1] $: for any $q, q^{'}\in Q $
and any $\sigma\in\Sigma$,
\begin{equation}
\delta(q, \sigma)=q^{'}   \hskip 8mm  {\it iff}  \hskip 8mm
\eta(q, \sigma, q^{'})>0.
\end{equation}
And $\delta$ can be extended to $\Sigma^{*}$ in the usual manner,
i.e., for any $q\in Q $, $ s\in\Sigma^{*}$ and $\sigma\in\Sigma$,
$$\delta(q, \epsilon)=q, \hskip 8mm \delta(q, s\sigma)=\delta(\delta(q, s), \sigma).$$
It can be readily verified that $L(G)=L(G^{'})$.

We now present the construction of the codiagnoser for SDESs,
which is a DFA built on a given stochastic automaton $G=(Q,\Sigma,
\eta, q_{0})$ with some local observations. Without loss of
generality, assume there are two local projections
$P_{i}:\Sigma^{*}\rightarrow \Sigma_{o,i}^{*}$, where $i=1,2$. We
construct the codiagnoser for SDESs in terms of the following
steps.

{\it Step 1: Construct a diagnoser $G^{'}_{D}$ for the DFA $G^{'}$
deduced by $G$.}

Let $G^{'}=(Q,\Sigma, \delta, q_{0})$ be the DFA deduced by $G$
according to Definition 5. From the global projection
$P:\Sigma^{*}\rightarrow \Sigma_{o}^{*}$, where
$\Sigma_{o}=\Sigma_{o,1}\cup\Sigma_{o,2}$, we can construct a
diagnoser $G^{'}_{D}$ for $G^{'}$ by means of the approach in [25,
26], i.e.,
\begin{equation}
G^{'}_{D}=(Q_{D},\Sigma_{o}, \delta_{D}, \chi_{0}),
\end{equation}
where $Q_{D}$ is the set of states of the diagnoser,
$\Sigma_{o}=\Sigma_{o,1}\cup\Sigma_{o,2}$, \hskip 2mm$\delta_{D}$
is the transition function of the diagnoser, and the initial state
of the diagnoser is $\chi_{0}=\left\{(q_{0}, N)\right\}\in Q_{D}$.

{\it Step 2: Construct the local stochastic diagnosers
$\left\{G_{d}^{i}: i=1,2\right\}$ for $G$.}

According to the projections $P_{i}:\Sigma^{*}\rightarrow
\Sigma_{o,i}^{*}$, where $i=1,2$, we can construct two local
stochastic diagnosers $G_{d}^{1}$ and $ G_{d}^{2}$ by means of the
approach in [31], i.e.,
\begin{equation}
G_{d}^{1}=(Q_{1}, \Sigma_{o,1}, \delta_{1}, \chi_{0}, \Phi_{1},
\phi_{0}),
\end{equation}
\begin{equation}
G_{d}^{2}=(Q_{2}, \Sigma_{o,2}, \delta_{2}, \chi_{0}, \Phi_{2},
\phi_{0}),
\end{equation}
where $\Phi_{1}$ and $\Phi_{2}$ are the sets of probability
transition matrices, and $\phi_{0}$ is the initial probability
mass function on $\chi_{0}$, and each element $q^{1}_{i}\in Q_{1}$
or $q^{2}_{i}\in Q_{2}$ is of the form
\begin{equation}
q^{1}_{i}=\left\{(q^{1}_{i1}, \ell^{1}_{i1}), \cdots, (q^{1}_{im},
\ell^{1}_{im}) \right\}, \hskip 10mm q^{2}_{i}=\left\{(q^{2}_{i1},
\ell^{2}_{i1}), \cdots, (q^{2}_{in}, \ell^{2}_{in}) \right\},
\end{equation}
where $q^{1}_{ij}, q^{2}_{ij}\in Q$ and $\ell^{1}_{ij},
\ell^{2}_{ij}\in \triangle=\left\{N\right\}\cup 2^{\left\{F_{1},
\cdots, F_{k}\right\}}$. We can refer to [31] for the details.

{\it Step 3: Construct the codiagnoser $G_{T}$ of testing the
codiagnosability for $G$.}

Although system $G$ is a stochastic automaton, the codiagnoser
that we will construct subsequently to test the codiagnosability
is a DFA, which is interpreted as follows. On the one side, the
local diagnosers $G_{d}^{1}$ and $G_{d}^{2}$ are stochastic, so
the codiagnoser being DFA has also appended a probabilistic
structure through $G_{d}^{1}$ and $G_{d}^{2}$. On the other side,
we can decrease the cost of constructing the codiagnoser since a
DFA is simpler than a stochastic automaton.

The codiagnoser of testing the codiagnosability for $G$ is
constructed as a DFA
\begin{equation}
G_{T}=(Q_{T}, \Sigma_{T}, \delta_{T}, q^{T}_{0}),
\end{equation}
where $Q_{T}$ is the set of states of the codiagnoser,
$\Sigma_{T}$ is the set of inputting events, $\delta_{T}$ is the
transition function, and $q^{T}_{0}$ is initial element. More
specifically, they are defined as follows:

(1) $Q_{T}=Q_{D}\times Q_{1}\times Q_{2}$, and element
$q^{T}=(q^{D}, q^{1}, q^{2})\in Q_{T}$ is of the form
\begin{equation}
q^{T}=(q^{D}, \left\{ (q^{1}_{1}, \ell^{1}_{1}), \cdots,
(q^{1}_{m}, \ell^{1}_{m})\right\},  \left\{(q^{2}_{1},
\ell^{2}_{1}), \cdots, (q^{2}_{n}, \ell^{2}_{n})\right\}),
\end{equation}
where $q^{1}=\left\{ (q^{1}_{1}, \ell^{1}_{1}), \cdots,
(q^{1}_{m}, \ell^{1}_{m})\right\}\in Q_{1}$, and
$q^{2}=\left\{(q^{2}_{1}, \ell^{2}_{1}), \cdots, (q^{2}_{n},
\ell^{2}_{n})\right\}\in Q_{2}$. A triple $(q^{D}, (q^{1}_{i},
\ell^{1}_{i}), (q^{2}_{j}, \ell^{2}_{j}))$ is called a {\it
component of $q^{T}$}, where $(q^{1}_{i}, \ell^{1}_{i})\in q^{1}$
and $(q^{2}_{j}, \ell^{2}_{j}))\in q^{2}$.

(2) $\Sigma_{T}\subseteq
\Sigma_{o}\times\Sigma_{o}\times\Sigma_{o}$, and $\sigma^{T}\in
\Sigma_{T}$ is of the form $\sigma^{T}=(\sigma^{D}, \sigma^{1},
\sigma^{2})$, where
\begin{equation}
\sigma^{1}=\left\{
\begin{array}{ll}
\sigma^{D},    \hskip 3mm&{\rm if}  \hskip 3mm \sigma^{D}\in \Sigma_{o,1},\\
\epsilon, \hskip 3mm&{\rm if}  \hskip 3mm
\sigma^{D}\notin\Sigma_{o,1},
\end{array}
\right. \hskip 8mm \sigma^{2}=\left\{
\begin{array}{ll}
\sigma^{D},    \hskip 3mm&{\rm if}  \hskip 3mm \sigma^{D}\in \Sigma_{o,2},\\
\epsilon, \hskip 3mm&{\rm if}  \hskip 3mm
\sigma^{D}\notin\Sigma_{o,2}.
\end{array}
\right.
\end{equation}

(3) The transition function $\delta_{T}$ is defined as: for any
$q^{T}=(q^{D}, q^{1}, q^{2})\in Q_{T}$ and for any
$\sigma^{T}=(\sigma^{D}, \sigma^{1}, \sigma^{2})\in \Sigma_{T}$,
we discuss it by the following three cases.

i) If $\sigma^{D}\in \Sigma_{o,1}\cap\Sigma_{o,2}$, then
$\sigma^{T}=(\sigma^{D}, \sigma^{D}, \sigma^{D})$, and
\begin{equation}
\begin{array}{ll}
\delta_{T}(q^{T}, \sigma^{T})=(\delta_{D}(q^{D}, \sigma^{D}),
\hskip 2mm
\delta_{1}(q^{1}, \sigma^{D}), \hskip 2mm \delta_{2}(q^{2}, \sigma^{D}))\\
\Leftrightarrow \hskip 5mm  \delta_{D}(q^{D}, \sigma^{D})\neq
\emptyset,\hskip 3mm \delta_{1}(q^{1}, \sigma^{D})\neq
\emptyset,\hskip 3mm \delta_{2}(q^{2}, \sigma^{D})\neq \emptyset.
\end{array}
\end{equation}

ii) If $\sigma^{D}\in \Sigma_{o,1}-\Sigma_{o,2}$, then
$\sigma^{T}=(\sigma^{D}, \sigma^{D}, \epsilon)$, and
\begin{equation}
\begin{array}{ll}
\delta_{T}(q^{T}, \sigma^{T})=(\delta_{D}(q^{D}, \sigma^{D}),
\hskip 2mm
\delta_{1}(q^{1}, \sigma^{D}), \hskip 2mm q^{2})\\
\Leftrightarrow \hskip 5mm  \delta_{D}(q^{D}, \sigma^{D})\neq
\emptyset,\hskip 3mm \delta_{1}(q^{1}, \sigma^{D})\neq \emptyset.
\end{array}
\end{equation}

iii) If $\sigma^{D}\in \Sigma_{o,2}-\Sigma_{o,1}$, then
$\sigma^{T}=(\sigma^{D}, \epsilon, \sigma^{D})$, and
\begin{equation}
\begin{array}{ll}
\delta_{T}(q^{T}, \sigma^{T})=(\delta_{D}(q^{D}, \sigma^{D}),
\hskip 2mm
q^{1}, \hskip 2mm \delta_{2}(q^{2}, \sigma^{D}))\\
\Leftrightarrow \hskip 5mm  \delta_{D}(q^{D}, \sigma^{D})\neq
\emptyset,\hskip 3mm \delta_{2}(q^{2}, \sigma^{D})\neq \emptyset.
\end{array}
\end{equation}

(4) The initial element of $G_{T}$ is defined as
$q^{T}_{0}=\left\{ \chi_{0}, \chi_{0}, \chi_{0}\right\}\in Q_{T}$,
where $\chi_{0}=\left\{(q_{0}, N)\right\}$.

This codiagnoser associated with the local stochastic diagnosers
$G_{d}^{1}$ and $G_{d}^{2}$ (the part of dash line in Fig. 2) will
be used to perform decentralized failure diagnosis of SDESs and to
describe a necessary and sufficient condition of the
codiagnosability for SDESs in Section IV.

\setlength{\unitlength}{0.05in}

\begin{picture}(80,48)
\put(45,45){\framebox(6,4)[c]{$G$}}
\put(35,33){\framebox(6,4)[c]{$P_{1}$}}
\put(55,33){\framebox(6,4)[c]{$P_{2}$}}
\put(35,20){\framebox(6,5)[c]{$G_{d}^{1}$}}
\put(55,20){\framebox(6,5)[c]{$G_{d}^{2}$}}
\put(45,10){\framebox(6,4)[c]{$G_{T}$}}
\put(45,45){\vector(-1,-1){8}} \put(51,45){\vector(1,-1){8}}
\put(37,32.5){\vector(0,-1){7.5}}
\put(59,32.5){\vector(0,-1){7.5}} \put(37,19.5){\vector(1,-1){8}}
\put(59,19.5){\vector(-1,-1){8}} \put(28,8){\dashbox(40,20){}}

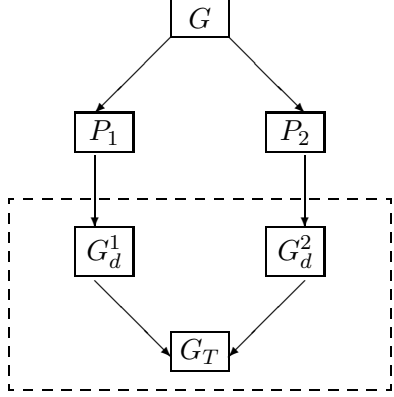
\put(45,0){\makebox(15,4)[c]{{\footnotesize Fig. 2. The
architecture of decentralized failure diagnosis of SDESs.}}}
\end{picture}

\section*{IV.  Necessary and Sufficient Condition of Codiagnosability for SDESs}
In this section, we give some properties of the codiagnoser. In
particular, a necessary and sufficient condition of the
codiagnosability for SDESs is presented. And we propose an
approach in detail to check whether codiagnosability is violated.
As well, some examples are given to illustrate the results we
present.

\subsection*{{\it A. Some Properties of the Codiagnoser}}
Firstly, we give some basic properties of the codiagnosers as
follows.

{\it Proposition 3:}  Let $G=(Q,\Sigma, \eta, q_{0})$ be a
stochastic automaton with two projections
$P_{i}:\Sigma^{*}\rightarrow \Sigma_{o,i}^{*}$  (where $i=1,2$).
$G_{T}=(Q_{T}, \Sigma_{T}, \delta_{T}, q^{T}_{0})$ is the
codiagnoser of $G$. For any $q^{T}=(q^{D}, q^{1}, q^{2})\in
Q_{T}$, there exists $s^{T}=(s^{D}, s^{1}, s^{2})\in
\Sigma_{T}^{*}$ such that
\begin{equation}
\delta_{D}(\chi_{0}, s^{D})=q^{D}, \hskip 4mm \delta_{1}(\chi_{0},
s^{1})=q^{1},  \hskip 4mm \delta_{2}(\chi_{0}, s^{2})=q^{2},
\end{equation}
\begin{equation}
P_{1}(s^{D})=P_{1}(s^{1}),  \hskip 8mm P_{2}(s^{D})=P_{2}(s^{2}),
\end{equation}
where $s^{D}\in \Sigma_{o}^{*}$, $s^{1}\in \Sigma_{o,1}^{*}$, and
$ s^{2}\in \Sigma_{o,2}^{*}$.

{\it Proof:} From the above construction of the codiagnoser, we
know that for any $q^{T}=(q^{D}, q^{1}, q^{2})\in Q_{T}$, there
exists $s^{T}\in \Sigma_{T}^{*}$ such that $\delta_{T}(q^{T}_{0},
s^{T})=q^{T}$. We prove the proposition by induction on $\parallel
s^{T}\parallel $, the length of $s^{T}$.

{\it Basis:}  If $\parallel s^{T}\parallel=1$, then from (31) we
have $s^{T}=(\sigma^{D}, \sigma^{D}, \sigma^{D})$, or
$s^{T}=(\sigma^{D}, \sigma^{D}, \epsilon)$, or $s^{T}=(\sigma^{D},
\epsilon, \sigma^{D})$. It is clear that Eqs. (35, 36) hold.

{\it Induction:} Let $\delta_{T}(q^{T}_{1}, \sigma^{T})=q^{T}$
where $q^{T}_{1}=(q^{D}_{1}, q^{1}_{1}, q^{2}_{1})\in Q_{T}$. By
the assumption of induction there exists $s^{T}\in
\Sigma_{T}^{*}$, where $\parallel s^{T}\parallel=n$, such that
$\delta_{T}(q^{T}_{0}, s^{T})=q^{T}_{1}$,  and Eqs. (35, 36) hold.
There are three cases to be considered for $s^{T}\sigma^{T}$,
where $\sigma^{T}=(\sigma^{D}, \sigma^{1}, \sigma^{2})$.

{\it Case 1:} If $\sigma^{D}\in \Sigma_{o,1}\cap\Sigma_{o,2}$,
then $\sigma^{D}=\sigma^{1}=\sigma^{2}$. Therefore,
$$\delta_{T}(q^{T}_{0}, s^{T}\sigma^{T})=\delta_{T}(\delta_{T}(q^{T}_{0},
s^{T}), \sigma^{T})=\delta_{T}(q^{T}_{1}, \sigma^{T})=q^{T}.$$
That is,
$$\delta_{D}(\chi_{0}, s^{D}\sigma^{D})=q^{D}, \hskip 4mm
\delta_{1}(\chi_{0}, s^{1}\sigma^{1})=q^{1}, \hskip 4mm
\delta_{2}(\chi_{0}, s^{2}\sigma^{2})=q^{2},$$
$$P_{1}(s^{D}\sigma^{D})=P_{1}(s^{D})P_{1}(\sigma^{D})
=P_{1}(s^{1})P_{1}(\sigma^{1})=P_{1}(s^{1}\sigma^{1}).$$
 Similarly, we have $P_{2}(s^{D}\sigma^{D})=P_{2}(s^{2}\sigma^{2})$.

{\it Case 2:} If $\sigma^{D}\in \Sigma_{o,1}-\Sigma_{o,2}$, then
$\sigma^{D}=\sigma^{1}$, but $\sigma^{2}=\epsilon$. Therefore,
$$\delta_{1}(\chi_{0}, s^{1}\sigma^{1})=\delta_{1}(\delta_{1}(\chi_{0},
s^{1}), \sigma^{1})=\delta_{1}(q^{1}_{1}, \sigma^{1})=q^{1},$$ but
$\delta_{2}(\chi_{0}, s^{2}\sigma^{2})=\delta_{2}(\chi_{0},
s^{2})=q^{2}_{1}=q^{2}$. We also have
$P_{1}(s^{D}\sigma^{D})=P_{1}(s^{1}\sigma^{1})$ for the same
reason as in Case 1, and
$$P_{2}(s^{D}\sigma^{D})=P_{2}(s^{D})P_{2}(\sigma^{D})
=P_{2}(s^{2})\epsilon=P_{2}(s^{2}\sigma^{2}).$$

{\it Case 3:} If $\sigma^{D}\in \Sigma_{o,2}-\Sigma_{o,1}$, then
$\sigma^{D}=\sigma^{2}$, but $\sigma^{1}=\epsilon$, we can
similarly verify that Eqs. (35, 36) hold for $s^{T}\sigma^{T}$.
\hfill $\Box$

{\it Proposition 4:} Let $G=(Q,\Sigma, \eta, q_{0})$ be a
stochastic automaton with two projections
$P_{i}:\Sigma^{*}\rightarrow \Sigma_{o,i}^{*}$  (where $i=1,2$)
and let $G_{T}=(Q_{T}, \Sigma_{T}, \delta_{T}, q^{T}_{0})$ be the
codiagnoser of $G$, and $q^{T}=(q^{D}, q^{1}, q^{2})\in Q_{T}$.

(1) If $(q^{1}_{a}, \ell^{1}_{a}), (q^{1}_{b}, \ell^{1}_{b})\in
q^{1}$, $F\in \ell^{1}_{a}$ but $F\notin \ell^{1}_{b}$, then there
exist $\omega_{1}, \omega_{2}\in \Sigma_{o,1}^{*}$ such that
$\delta(q_{0}, \omega_{1})=q^{1}_{a}$, $\delta(q_{0},
\omega_{2})=q^{1}_{b}$, $\sigma_{f}\in \omega_{1}$,
$\sigma_{f}\notin \omega_{2}$, and
$P_{1}(\omega_{1})=P_{1}(\omega_{2})$.

(2) If $(q^{2}_{a}, \ell^{2}_{a}), (q^{2}_{b}, \ell^{2}_{b})\in
q^{2}$, $F\in \ell^{2}_{a}$ but $F\notin \ell^{2}_{b}$, then there
exist $\upsilon_{1}, \upsilon_{2}\in \Sigma_{o,2}^{*}$ such that
$\delta(q_{0}, \upsilon_{1})=q^{2}_{a}$, $\delta(q_{0},
\upsilon_{2})=q^{2}_{b}$, $\sigma_{f}\in \upsilon_{1}$,
$\sigma_{f}\notin \upsilon_{2}$, and
$P_{2}(\upsilon_{1})=P_{2}(\upsilon_{2})$.

{\it Proof:} (1) Let $(q^{1}_{a}, \ell^{1}_{a}), (q^{1}_{b},
\ell^{1}_{b})\in q^{1}$, $F\in \ell^{1}_{a}$ but $F\notin
\ell^{1}_{b}$. Since every component in each element of $Q_{1}$ is
reachable from the initial state $q_{0}$. Therefore, there exist
$\omega_{1}, \omega_{2}\in \Sigma_{o,1}^{*}$ such that
$\delta(q_{0}, \omega_{1})=q^{1}_{a}$ and $\delta(q_{0},
\omega_{2})=q^{1}_{b}$. By the definition of label propagation
function [31], we have $\sigma_{f}\in \omega_{1}$ and
$\sigma_{f}\notin \omega_{2}$, because $F\in \ell^{1}_{a}$ and
$F\notin \ell^{1}_{b}$. From the construction of diagnoser
$G_{d}^{1}$, due to both $(q^{1}_{a}, \ell^{1}_{a})$ and
$(q^{1}_{b}, \ell^{1}_{b})$ in $q^{1}$, we know that $\omega_{1}$
and $\omega_{2}$ have the same strings filtered by projection
$P_{1}$. That is, $P_{1}(\omega_{1})=P_{1}(\omega_{2})$.

(2) It can be proved similarly. \hfill $\Box$

\subsection*{{\it B. Necessary and Sufficient Condition of Codiagnosability for SDESs}}
In this subsection, we will present the necessary and sufficient
condition of codiagnosability for SDESs.

{\it Definition 6:} Let $G_{T}=(Q_{T}, \Sigma_{T}, \delta_{T},
q^{T}_{0})$ be a codiagnoser of a stochastic automaton
$G=(Q,\Sigma,\eta, q_{0})$. A set $\left\{q^{T}_{1},
\sigma^{T}_{1}, q^{T}_{2}, \sigma^{T}_{2}, \ldots, q^{T}_{k},
\sigma^{T}_{k}, q^{T}_{1}\right\}$ is said to form a {\it cycle}
in $G_{T}$, if
$$\delta_{T}(q^{T}_{j},
\sigma^{T}_{j})=q^{T}_{j+1}, \hskip 8mm\delta_{T}(q^{T}_{k},
\sigma^{T}_{k})=q^{T}_{1},$$ where $q^{T}_{1}, q^{T}_{2},\ldots,
q^{T}_{k}\in Q_{T}$,\hskip 3mm $\sigma^{T}_{1},
\sigma^{T}_{2},\ldots,\sigma^{T}_{k}\in \Sigma_{T}$, and
$j=1,2,\ldots,k-1$.

{\it Definition 7:}  Let $G_{T}=(Q_{T}, \Sigma_{T}, \delta_{T},
q^{T}_{0})$ be a codiagnoser of a stochastic automaton
$G=(Q,\Sigma,\eta, q_{0})$, and $q^{T}=(q^{D}, q^{1}, q^{2})\in
Q_{T}$.

(1) If both $q^{1}$ and $q^{2}$ are $F-$certain in diagnoser
$G_{d}^{1}$ and diagnoser $G_{d}^{2}$, respectively, then $q^{T}$
is called to be {\it $F-$certain} in $G_{T}$.

(2) If both $q^{1}$ and $q^{2}$ are $F-$uncertain in $G_{d}^{1}$
and $G_{d}^{2}$, respectively, then $q^{T}$ is called to be {\it
$F-$uncertain} in $G_{T}$.

For example, in Fig. 9,  the states $(\left\{(1,N)\right\},
\left\{(0,N)\right\}, \left\{(1,N)\right\})$ and\\
$(\left\{(6,F)\right\}, \left\{(6,F)\right\}, \left\{(5,F),
(6,F)\right\})$ are $F-$certain in $G_{T}$, the state
$$(\left\{(2,F), (3,F), (4,N)\right\}, \left\{(2,F), (3,F), (4,N),
(5,F)\right\}, \left\{(2,F), (3,F), (4,N)\right\})$$ is
$F-$uncertain in $G_{T}$, but  $(\left\{(5,F)\right\},
\left\{(2,F), (3,F), (4,N), (5,F)\right\}, \left\{(5,F),
(6,F)\right\})$ is neither $F-$certain nor $F-$uncertain in
$G_{T}$.

{\it Definition 8:}  Let $G_{T}=(Q_{T}, \Sigma_{T}, \delta_{T},
q^{T}_{0})$ be a codiagnoser of a stochastic automaton
$G=(Q,\Sigma,\eta, q_{0})$.  Let $q^{T}=(q^{D}, q^{1}, q^{2})\in
Q_{T}$, and, let $(q^{D}, (q^{1}_{a}, \ell^{1}_{a}), (q^{2}_{b},
\ell^{2}_{b}))$ be a component of $q^{T}$. If both $(q^{1},
q^{1}_{a}, \ell^{1}_{a})$ and $(q^{2}, q^{2}_{b}, \ell^{2}_{b})$
are recurrent components of $q^{1}$ in $G_{d}^{1}$ and $q^{2}$ in
$G_{d}^{2}$, respectively, then $(q^{D}, (q^{1}_{a},
\ell^{1}_{a}), (q^{2}_{b}, \ell^{2}_{b}))$ is called {\it a
recurrent component of $q^{T}$ in $G_{T}$}. Furthermore, if the
recurrent component $(q^{D}, (q^{1}_{a}, \ell^{1}_{a}),
(q^{2}_{b}, \ell^{2}_{b}))$ satisfies $F\in \ell^{1}_{a}$ and
$F\in \ell^{2}_{b}$, then it is called {\it a recurrent component
bearing the label $F$}.

{\it Definition 9:}  Let $G_{T}=(Q_{T}, \Sigma_{T}, \delta_{T},
q^{T}_{0})$ be a codiagnoser of a stochastic automaton
$G=(Q,\Sigma,\eta, q_{0})$.  Let $q^{T}=(q^{D}, q^{1}, q^{2})\in
Q_{T}$, and, let $(q^{D}, (q^{1}_{a}, \ell^{1}_{a}), (q^{2}_{b},
\ell^{2}_{b}))$ be a component of $q^{T}$. If
$q^{1}_{a}=q^{2}_{b}$, $\ell^{1}_{a}=\ell^{2}_{b}$, and there
exists $\omega\in L$ such that
\begin{equation}
\delta(q_{0}, \omega)=q^{1}_{a}, \hskip 4mm \delta_{1}(\chi_{0},
P_{1}(\omega))=q^{1},  \hskip 4mm \delta_{2}(\chi_{0},
P_{2}(\omega))=q^{2},
\end{equation}
then the component $(q^{D}, (q^{1}_{a}, \ell^{1}_{a}), (q^{2}_{b},
\ell^{2}_{b}))$ is called {\it uniform}.

For example, in Fig. 5, $(\left\{(3,F)\right\}, (3,F), (3,F))$ is
a uniform component of state $q^{T}$ in codiagnoser $G_{T}$, where
$q^{T}=(\left\{(3,F)\right\}, \left\{(2,F), (3,F), (4,N),
(5,F)\right\}, \left\{(3,F)\right\})$. In fact, we can take
$\omega=d\sigma_{f}ca\in L$ which satisfies (37).

{\it Proposition 5:} Let $G_{T}=(Q_{T}, \Sigma_{T}, \delta_{T},
q^{T}_{0})$ be a codiagnoser of stochastic automaton $G=(Q,\Sigma,
\eta, q_{0})$ with two projections $P_{i}:\Sigma^{*}\rightarrow
\Sigma_{o,i}^{*}$, where $i=1,2$. If $L$ is not codiagnosable with
respect to $\left\{P_{i}: i=1, 2\right\}$, then there exists an
$F-$uncertain state with a recurrent component bearing the label
$F$ in $G_{T}$.

{\it Proof:} Since $L$ is not codiagnosable, by Proposition 2,
there exists $\epsilon>0$, and for any $ n_{i}\in {\bf N}$, there
exist $s\in \Psi(\Sigma_{f})$ and $ t\in L/s $ (where $ n\geq
n_{i}$), such that
\begin{equation}
Pr(t:D_{1}(st)=0 \mid t\in L/s \wedge
\parallel t
\parallel=n)\geq\epsilon,
\end{equation}
\begin{equation}
 Pr(t:D_{2}(st)=0 \mid t\in L/s \wedge
\parallel t
\parallel=n)\geq\epsilon.
\end{equation}
Denote by $\delta_{D}(\chi_{0}, P(st))=q^{D}, \hskip 4mm
\delta_{1}(\chi_{0}, P_{1}(st))=q^{1},  \hskip 4mm
\delta_{2}(\chi_{0}, P_{2}(st))=q^{2}$. We assert that $(q^{D},
q^{1}, q^{2})\in Q_{T}$ is an $F-$uncertain state in $G_{T}$.
Otherwise, if $q^{1}\in Q_{1}$ is an $F-$certain state in
$G_{d}^{1}$, then for any $\omega\in P_{1}^{-1}[P_{1}(st)]$, we
have $\sigma_{f}\in \omega$. That is, $D_{1}(st)=1$ always holds.
Therefore,
$$Pr(t:D_{1}(st)=0 \mid t\in L/s \wedge
\parallel t
\parallel=n)=0,$$
which is in contradiction with (38). So $q^{1}$ is an
$F-$uncertain state in $G_{d}^{1}$. Similarly, from (39) we can
know that $q^{2}$ is an $F-$uncertain state in $G_{d}^{2}$. Hence,
$(q^{D}, q^{1}, q^{2})\in Q_{T}$ is an $F-$uncertain state in
$G_{T}$. Furthermore, by Lemma 1 and (38, 39), both $q^{1}$ and
$q^{2}$ contain respectively a recurrent component bearing the
label $F$. That is, $(q^{D}, q^{1}, q^{2})\in Q_{T}$ is an
$F-$uncertain state in $G_{T}$ with a recurrent component bearing
the label $F$. \hfill $\Box$

Using the above results, we present a necessary and sufficient
condition of the codiagnosability for SDESs as follows.

{\it Theorem 1:} Let $G=(Q,\Sigma,\eta, q_{0})$ be a stochastic
automaton with two local projections $P_{i}:\Sigma^{*}\rightarrow
\Sigma_{o,i}^{*}$, where $i=1, 2$. Let $G_{T}=(Q_{T}, \Sigma_{T},
\delta_{T}, q^{T}_{0})$ be a codiagnoser of $G$. Then $L=L(G)$ is
not codiagnosable with respect to $\left\{P_{i}\right\}$, if and
only if, there exists a cycle $C^{T}=\left\{q^{T}_{k},
\sigma^{T}_{k}, q^{T}_{k+1}, \sigma^{T}_{k+1}, \ldots, q^{T}_{k
+h}, \sigma^{T}_{k+h}, q^{T}_{k}\right\}$ in $G_{T}$ such that
each state $q^{T}_{k+i}$ with a uniform recurrent component
bearing the label $F$ is $F-$uncertain in $G_{T}$, where $i\in [0,
h]$.

{\it Proof: Sufficiency: } Assume that there exists a cycle
$$C^{T}=\left\{q^{T}_{k}, \sigma^{T}_{k}, q^{T}_{k+1},
\sigma^{T}_{k+1}, \ldots, q^{T}_{k+h}, \sigma^{T}_{k+h},
q^{T}_{k}\right\}$$ in $G_{T}$, such that each state
$q^{T}_{k+i}=(q^{D}_{k+i}, q^{1}_{k+i}, q^{2}_{k+i})$ with a
uniform recurrent component (denoted by $(q^{D}_{k+i}, (q_{a},
\ell_{a}), (q_{a}, \ell_{a}))$) bearing the label $F$ is
$F-$uncertain in $G_{T}$, where $i\in [0, h]$. By Definition 9,
there exists $\omega\in L$ such that $\sigma_{f}\in \omega$ and
\begin{equation}
\delta(q_{0}, \omega)=q_{a}, \hskip 6mm \delta_{1}(\chi_{0},
P_{1}(\omega))=q^{1}_{k+i},  \hskip 6mm \delta_{2}(\chi_{0},
P_{2}(\omega))=q^{2}_{k+i}.
\end{equation}
Since $q^{T}_{k+i}$ is a state of the cycle $C^{T}$, there is a
path ending with the cycle $C^{T}$ in $G_{T}$:
\begin{equation}
path=\sigma^{T}_{0}\sigma^{T}_{1}
\cdots(\sigma^{T}_{k}\cdots\sigma^{T}_{k+i}\cdots\sigma^{T}_{k+h})^{n}.
\end{equation}

We take $s\in \overline{\omega}$, $t\in L/s $ and $u\in L/st $
such that $s\in \Psi(\Sigma_{f})$, $st=\omega$ and
$$P_{1}(stu)=\sigma^{1}_{0}\sigma^{1}_{1}
\cdots(\sigma^{1}_{k}\cdots\sigma^{1}_{k+i}\cdots\sigma^{1}_{k+h})^{n},$$
$$P_{2}(stu)=\sigma^{2}_{0}\sigma^{2}_{1}
\cdots(\sigma^{2}_{k}\cdots\sigma^{2}_{k+i}\cdots\sigma^{2}_{k+h})^{n}.$$
Notice that $q^{T}_{k+i}$ is $F-$uncertain in $G_{T}$, there exist
$\omega_{1}\in P_{1}^{-1}[P_{1}(st)]$ and $\omega_{2}\in
P_{2}^{-1}[P_{2}(st)]$ such that $\sigma_{f}\notin \omega_{1}$ and
$\sigma_{f}\notin \omega_{2}$, i.e., $D_{1}(st)=0$ and
$D_{2}(st)=0$. That is,
\begin{equation}
Pr(u: D_{1}(stu)=0 \mid tu\in L/s \wedge
\parallel tu
\parallel=n)=1,
\end{equation}
\begin{equation}
Pr(u: D_{2}(stu)=0 \mid tu\in L/s \wedge
\parallel tu
\parallel=n)=1.
\end{equation}
Furthermore, $Pr(t: t\in L/s)>0$ since $t\in L/s$. Let
$0<\epsilon<Pr(t: t\in L/s)$, then by (42, 43) we have
\begin{equation}
\begin{array}{ll}
Pr(tu: D_{1}(stu)=0 \mid tu\in L/s \wedge
\parallel tu
\parallel=n)\\
=Pr(t: t\in L/s)Pr(u: D_{1}(stu)=0 \mid tu\in L/s \wedge
\parallel tu
\parallel=n)>\epsilon,
\end{array}
\end{equation}
and
\begin{equation}
\begin{array}{ll}
Pr(tu: D_{2}(stu)=0 \mid tu\in L/s \wedge
\parallel tu
\parallel=n)\\
=Pr(t: t\in L/s)Pr(u: D_{2}(stu)=0 \mid tu\in L/s \wedge
\parallel tu
\parallel=n)>\epsilon.
\end{array}
\end{equation}
By Remark 3, we obtain that $L$ is not codiagnosable with respect
to $\left\{P_{i}: i=1, 2\right\}$.

{\it Necessity: } Assume that $L$ is not codiagnosable. By
Proposition 2 and Remark 3, there is $\epsilon>0$, such that for
any $n_{i}\in {\bf N}$, there exist $s\in \Psi(\Sigma_{f_{i}})$
and $ t\in L/s $, where $ \parallel t
\parallel=n\geq n_{i}$, satisfying  Ineqs. (22, 23).
Therefore, there exist $\omega_{1}\in P_{1}^{-1}[P_{1}(st)]$ and
$\omega_{2}\in P_{2}^{-1}[P_{2}(st)]$ such that $\sigma_{f}\notin
\omega_{1}$ and $\sigma_{f}\notin \omega_{2}$. Let $\parallel Q
\parallel$ be the number of states of $G$. If we take $n_{i}$ big enough such that $\parallel
st
\parallel > \parallel Q \parallel $, then there will be a cycle
$C$ in $G$ along string $st$. According to Assumption ({\it A2}),
the cycle $C$ must contain an observable event in $\Sigma_{o}$.
Therefore, there exists a cycle $C^{D}$ in diagnoser $G^{'}_{D}$
corresponding to the cycle $C$, denoted by
$$C^{D}=\left\{q^{D}_{k}, \sigma^{D}_{k}, q^{D}_{k+1},
\sigma^{D}_{k+1}, \ldots, q^{D}_{k+h}, \sigma^{D}_{k+h},
q^{D}_{k}\right\}.$$ We denote the path in $G^{'}_{D}$ ending with
the cycle $C^{D}$ as
$$path=\sigma^{D}_{0}\sigma^{D}_{1}
\cdots(\sigma^{D}_{k}\cdots\sigma^{D}_{k+i}\cdots\sigma^{D}_{k+h})^{n}.$$
According to the construction of the codiagnoser $G_{T}$, we can
obtain a cycle in $G_{T}$:
\begin{equation}
C^{T}=\left\{q^{T}_{k}, \sigma^{T}_{k}, q^{T}_{k+1},
\sigma^{T}_{k+1}, \ldots, q^{T}_{k+h}, \sigma^{T}_{k+h},
q^{T}_{k}\right\},
\end{equation}
where $\sigma^{T}_{j}=(\sigma^{D}_{j}, \sigma^{1}_{j},
\sigma^{2}_{j})$, $j\in [k,k+h]$. Denote
$$q^{1}_{i_{0}}=\delta_{1}(\chi_{0}, P_{1}(st)),\hskip 8mm
q^{2}_{i_{0}}=\delta_{2}(\chi_{0}, P_{2}(st)).$$

Notice that $st$ satisfies Ineq. (22, 23). By Lemma 1, both
$q^{1}_{i_{0}}$ and $q^{2}_{i_{0}}$ contain respectively a
recurrent component bearing the label $F$. Furthermore, from
$\sigma_{f}\in s $, $\sigma_{f}\notin \omega_{1}$ and
$\sigma_{f}\notin \omega_{2}$, we know that both $q^{1}_{i_{0}}$
and $q^{2}_{i_{0}}$ are $F-$uncertain states in $G_{d}^{1}$ and in
$G_{d}^{2}$, respectively. That is, $q^{T}_{i_{0}}=(q^{D}_{i_{0}},
q^{1}_{i_{0}}, q^{2}_{i_{0}})\in Q_{T}$ is an $F-$uncertain state
in $G_{T}$.

In the following we will verify that each state
$q^{T}_{j}=(q^{D}_{j}, q^{1}_{j}, q^{2}_{j})$ with a uniform
recurrent component (denoted by $(q^{D}_{j}, (q_{a}, \ell_{a}),
(q_{a}, \ell_{a}))$) bearing the label $F$ in the cycle $C^{T}$ is
$F-$uncertain in $G_{T}$, where $q^{T}_{j}\in Q_{T}$ and $j\in [k,
k+h]$.

{\it Case 1:} If $q^{T}_{j}=q^{T}_{i_{0}}$, then it has been
verified above that $q^{T}_{i_{0}}$ is $F-$uncertain in $G_{T}$.

{\it Case 2:} Assume $q^{T}_{j}\neq q^{T}_{i_{0}}$. Due to both
$q^{T}_{i_{0}}$ and $q^{T}_{j}$ being states in the cycle $C^{T}$,
there exists a component of $q^{1}_{j}$ reachable from the
recurrent component of $q^{1}_{i_{0}}$, and there exists a
component of $q^{2}_{j}$ reachable from the recurrent component of
$q^{2}_{i_{0}}$. Likewise, there exists a component of
$q^{1}_{i_{0}}$ reachable from the recurrent component
$(q^{1}_{j}, q_{a}, \ell_{a})$, and there exists a component of
$q^{2}_{i_{0}}$ reachable from the recurrent component
$(q^{2}_{j}, q_{a}, \ell_{a})$. By Lemma 2, both $q^{1}_{j}$ and
$q^{2}_{j}$ are $F-$uncertain in $G_{d}^{1}$ and in $G_{d}^{2}$,
respectively, since  $q^{T}_{i_{0}}$ is $F-$uncertain in $G_{T}$.
Therefore, $q^{T}_{j}=(q^{D}_{j}, q^{1}_{j}, q^{2}_{j})\in Q_{T}$
is an $F-$uncertain state in $G_{T}$. \hfill $\Box$

{\it Remark 4:} From the proof of Theorem 1, we know that if there
is only one projection in system $G$, then Theorem 1 degenerates
to Theorem 3 of [31]. Therefore, the centralized failure diagnosis
in [31] can be regarded as a special case of the decentralized
failure diagnosis here.

\subsection*{{\it C. The Computing Process of Checking the Codiagnosability in SDESs}}
Let $G=(Q,\Sigma,\eta, q_{0})$ be a stochastic automaton
 with two projections
$P_{i}:\Sigma^{*}\rightarrow \Sigma_{o,i}^{*}$, where $i=1, 2$.
$G^{'}$ is a DFA deduced by $G$. We give a computing process to
check whether the codiagnosability condition of Theorem 1 is
violated.

{\it Step 1: Construct the diagnoser $G^{'}_{D}$ (i.e., Eq. (25))
for the DFA $G^{'}$ and the local stochastic diagnosers
$\left\{G_{d}^{i}: i=1,2\right\}$ (i.e., Eqs. (26,27)) for $G$.}

This specific procedure can be seen in Part B of Section III for
the details.

{\it Step 2: Construct the codiagnoser $G_{T}=(Q_{T}, \Sigma_{T},
\delta_{T}, q^{T}_{0})$ (i.e., Eq. (29)) for $G$.}

Also, this specific procedure can be seen in Part B of Section III
for the details.

{\it Step 3: Check whether there exists a cycle in the codiagnoser
$G_{T}$.}

If there does not exist a cycle in the codiagnoser $G_{T}$, then
$L$ is codiagnosable with respect to $\left\{P_{i}\right\}$.
Otherwise, we perform the next step.

{\it Step 4: Check whether the states in each cycle satisfy the
following condition: each state with a uniform recurrent component
bearing the label $F$ is $F-$uncertain in $G_{T}$.}

If each state with a uniform recurrent component bearing the label
$F$ in each cycle is $F-$uncertain in $G_{T}$, then we further
perform the next step. Otherwise, $L$ is codiagnosable.

{\it Step 5: Check whether the codiagnosability is violated.}

 If there exists a cycle $C^{T}$ whose each state
$q^{T}$ with a uniform recurrent component bearing the label $F$
is $F-$uncertain, then $L$ is not codiagnosable. Otherwise, $L$ is
codiagnosable.

\subsection*{{\it D. Examples of  Codiagnosability for SDESs}}
In this subsection, we will give some examples to illustrate the
applications of the necessary and sufficient condition for the
codiagnosability of FDESs and its computing method we presented
above.

{\it Example 3.} Consider the stochastic automaton $G$ as in
Example 1. $L=L(G)$. Assume there are two projections
$P_{i}:\Sigma^{*}\rightarrow \Sigma_{o,i}^{*}$, where $i=1, 2$ and
$\Sigma_{o, 1}=\{a, b\}$, $\Sigma_{o, 2}=\{a, c\}$.

From Example 2, we know that $L$ is neither diagnosable with
respect to $P_{1}$, nor diagnosable with respect to $P_{2}$. But
Example 1 shows that $L$ is codiagnosable. In the following we use
Theorem 1 and the above computing process to test these results.

According to the global projection $P:\Sigma^{*}\rightarrow
\Sigma_{o}^{*}$ where
$\Sigma_{o}=\Sigma_{o,1}\cup\Sigma_{o,2}=\{a, b, c\}$, and the
projections $P_{i}:\Sigma^{*}\rightarrow \Sigma_{o,i}^{*}$, (where
$i=1,2$), we can construct the diagnoser
$G^{'}_{D}=(Q_{D},\Sigma_{o}, \delta_{D}, \chi_{0})$ for DFA
$G^{'}$ as in Fig. 3, and two local stochastic diagnosers
$G_{d}^{1}, G_{d}^{2}$ as in Fig. 4, where $G_{d}^{1}=(Q_{1},
\Sigma_{o,1}, \delta_{1}, \chi_{0}, \Phi_{1}, \phi_{0}), \hskip
3mm G_{d}^{2}=(Q_{2}, \Sigma_{o,2}, \delta_{2}, \chi_{0},
\Phi_{2}, \phi_{0})$.

\setlength{\unitlength}{0.05in}

\begin{picture}(150,43)
\put(0,20){\framebox(5,4)[c]{$0N$}}
\put(10,30){\framebox(5,4)[c]{$3F$}}
\put(10,10){\framebox(5,4)[c]{$6F$}}
\put(23,17){\framebox(5,9)[c]{$4N$}}
\put(23,23){\makebox(5,3)[c]{$2F$}}
\put(23,17){\makebox(5,3)[c]{$5F$}}

\put(-3,22){\vector(1,0){2.5}} \put(5,22){\vector(1,0){18}}
\put(2,20){\vector(1,-1){8}} \put(3,24){\vector(1,1){7}}
\put(23,19){\vector(-1,-1){8}} \put(23,24){\vector(-1,1){8}}
\put(13,36.8){\circle{6}} \put(15.8,37.3){\vector(0,-1){1}}
\put(13,7){\circle{6}} \put(15.8,7.5){\vector(0,-1){1}}
\put(31,22){\circle{6}} \put(31,19.2){\vector(-1,0){1}}

\put(4,16){\makebox(0,0)[c]{$b$}}
\put(5,28){\makebox(0,0)[c]{$c$}}
\put(13,23){\makebox(0,0)[c]{$a$}}
\put(21,14){\makebox(0,0)[c]{$b$}}
\put(21,29){\makebox(0,0)[c]{$c$}}
\put(17.5,7){\makebox(0,0)[c]{$a$}}
\put(17.5,37){\makebox(0,0)[c]{$a$}}
\put(30.3,17){\makebox(0,0)[c]{$a$}}

\put(10,0){\makebox(12,3)[c]{{\footnotesize Fig. 3. Diagnoser
$G^{'}_{D}$ in Example 3.}}}

\put(45,30){\framebox(5,4)[c]{$0N$}}
\put(53,10){\framebox(5,4)[c]{$6F$}}
\put(65,25){\framebox(5,12)[c]{}}

\put(65,31){\makebox(5,3)[c]{$3F$}}
\put(65,34){\makebox(5,3)[c]{$2F$}}
\put(65,28){\makebox(5,3)[c]{$4N$}}
\put(65,25){\makebox(5,3)[c]{$5F$}}

\put(40,32){\vector(1,0){5}} \put(50,32){\vector(1,0){15}}
\put(47,30){\vector(1,-2){8}} \put(66,25){\vector(-1,-1){11}}

\put(56,7){\circle{6}} \put(58.8,7.5){\vector(0,-1){1}}
\put(67,40){\circle{6}} \put(64.2,40){\vector(0,-1){1}}

\put(42,34){\makebox(0,0)[c]{$\phi_{0}$}}
\put(58,34){\makebox(0,0)[c]{$(a, \phi^{1}_{1})$}}
\put(58,40){\makebox(0,0)[c]{$(a, \phi^{1}_{4})$}}
\put(55,23){\makebox(0,0)[c]{$(b, \phi^{1}_{2})$}}
\put(65,18){\makebox(0,0)[c]{$(b, \phi^{1}_{5})$}}
\put(64,7){\makebox(0,0)[c]{$(a, \phi^{1}_{3})$}}

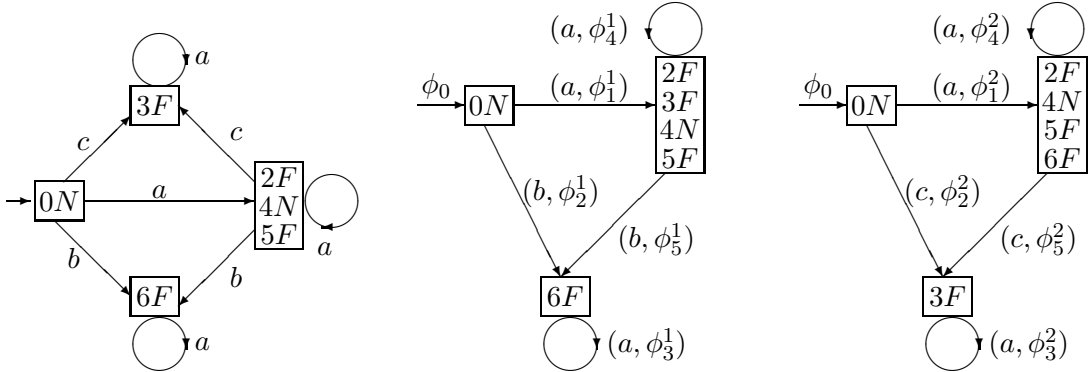
\put(65,0){\makebox(30,3)[c]{{\footnotesize Fig. 4. Local
diagnosers $G^{1}_{d}$ (left one) and $G^{2}_{d}$ (right one) in
Example 3.}}}

\put(85,30){\framebox(5,4)[c]{$0N$}}
\put(93,10){\framebox(5,4)[c]{$3F$}}
\put(105,25){\framebox(5,12)[c]{}}

\put(105,31){\makebox(5,3)[c]{$4N$}}
\put(105,34){\makebox(5,3)[c]{$2F$}}
\put(105,28){\makebox(5,3)[c]{$5F$}}
\put(105,25){\makebox(5,3)[c]{$6F$}}

\put(80,32){\vector(1,0){5}} \put(90,32){\vector(1,0){15}}
\put(87,30){\vector(1,-2){8}} \put(106,25){\vector(-1,-1){11}}

\put(96,7){\circle{6}} \put(98.8,7.5){\vector(0,-1){1}}
\put(107,40){\circle{6}} \put(104.2,40){\vector(0,-1){1}}

\put(82,34){\makebox(0,0)[c]{$\phi_{0}$}}
\put(98,34){\makebox(0,0)[c]{$(a, \phi^{2}_{1})$}}
\put(98,40){\makebox(0,0)[c]{$(a, \phi^{2}_{4})$}}
\put(95,23){\makebox(0,0)[c]{$(c, \phi^{2}_{2})$}}
\put(105,18){\makebox(0,0)[c]{$(c, \phi^{2}_{5})$}}
\put(104,7){\makebox(0,0)[c]{$(a, \phi^{2}_{3})$}}

\end{picture}

For the local stochastic diagnoser $G_{d}^{1}$, the set of
probability transition matrices $\Phi_{1}=\left\{\phi_{0},
\phi^{1}_{1}, \phi^{1}_{2}, \phi^{1}_{3}, \phi^{1}_{4},
\phi^{1}_{5} \right\}$, where $\phi_{0}=[1], \hskip 3mm
\phi^{1}_{1}=[0.35, 0.15, 0.2, 0.24], \hskip 3mm
\phi^{1}_{2}=[0.06], \hskip 3mm \phi^{1}_{3}=[1]$,

$$\phi^{1}_{4}= \left[
\begin{array}{cccc}
0.7 & 0.3& 0 & 0\\
0& 1 & 0& 0\\
0& 0& 1& 0\\
0 & 0& 0 & 0.8
\end{array}
\right], \hskip 3mm \phi^{1}_{5}=\left[
\begin{array}{c}
0\\
0\\
0\\
0.2
\end{array}
\right].
$$
Therefore, the recurrent components bearing $F$ are $(q^{1}_{2},
6, F)$ and $(q^{1}_{3}, 3, F)$, where
$q^{1}_{2}=\left\{(6,F)\right\}$ and $q^{1}_{3}=\left\{(2,F),
(3,F), (4,N), (5,F)\right\}$. Notice that $q^{1}_{3}$ is not
$F-$certain in $G_{d}^{1}$, so, neither is $L$ diagnosable with
respect to $P_{1}$ by Lemma 3.

For the local stochastic diagnoser $G_{d}^{2}$, the set of
probability transition matrices $\Phi_{2}=\left\{\phi_{0},
\phi^{2}_{1}, \phi^{2}_{2}, \phi^{2}_{3}, \phi^{2}_{4},
\phi^{2}_{5} \right\}$, where $\phi_{0}=[1], \hskip 3mm
\phi^{2}_{1}=[0.35, 0.2, 0.24, 0.06], \hskip 3mm
\phi^{2}_{2}=[0.15], \hskip 3mm \phi^{2}_{3}=[1]$,
$$\phi^{2}_{4}= \left[
\begin{array}{cccc}
0.7 & 0& 0 & 0\\
0& 1 & 0& 0\\
0& 0& 0.8& 0.2\\
0 & 0& 0 & 1
\end{array}
\right], \hskip 3mm \phi^{2}_{5}=\left[
\begin{array}{c}
0.3\\
0\\
0\\
0
\end{array}
\right].
$$
Therefore, the recurrent component bearing $F$ are $(q^{2}_{2}, 3,
F)$ and $(q^{2}_{3}, 6, F)$, where
$q^{2}_{2}=\left\{(3,F)\right\}$ and $q^{2}_{3}=\left\{(2,F),
(4,N), (5,F), (6,F)\right\})$. Notice that $q^{2}_{3}$ is not
$F-$certain in $G_{d}^{2}$, so $L$ is not diagnosable with respect
to $P_{2}$ by Lemma 3, either.

Now we construct the codiagnoser $G_{T}$ to test the
codiagnosability for $G$. The codiagnoser is a DFA $G_{T}=(Q_{T},
\Sigma_{T}, \delta_{T}, q^{T}_{0})$ as in Fig. 5, where
$\Sigma_{T}=\left\{(a, a, a), (b, b, \epsilon), (c, \epsilon,
c)\right\}$.

\setlength{\unitlength}{0.05in}

\begin{picture}(150,48)
\put(5,25){\framebox(15,4)[c]{}} \put(10,25){\line(0,1){4}}
\put(15,25){\line(0,1){4}} \put(6,25){\makebox(4,3)[c]{$0N$}}
\put(11,25){\makebox(4,3)[c]{$0N$}}
\put(16,25){\makebox(4,3)[c]{$0N$}}

\put(45,40){\framebox(15,4)[c]{}} \put(50,40){\line(0,1){4}}
\put(55,40){\line(0,1){4}} \put(46,40){\makebox(4,3)[c]{$3F$}}
\put(51,40){\makebox(4,3)[c]{$0N$}}
\put(56,40){\makebox(4,3)[c]{$3F$}}

\put(45,10){\framebox(15,4)[c]{}} \put(50,10){\line(0,1){4}}
\put(55,10){\line(0,1){4}} \put(46,10){\makebox(4,3)[c]{$6F$}}
\put(51,10){\makebox(4,3)[c]{$6F$}}
\put(56,10){\makebox(4,3)[c]{$0N$}}

\put(45,20){\framebox(15,15)[c]{}} \put(50,20){\line(0,1){15}}
\put(55,20){\line(0,1){15}} \put(46,29){\makebox(4,3)[c]{$2F$}}
\put(46,25){\makebox(4,3)[c]{$4N$}}
\put(46,20){\makebox(4,3)[c]{$5F$}}
\put(51,29){\makebox(4,3)[c]{$2F$}}
\put(51,26){\makebox(4,3)[c]{$3F$}}
\put(51,23){\makebox(4,3)[c]{$4N$}}
\put(51,20){\makebox(4,3)[c]{$5F$}}
\put(56,29){\makebox(4,3)[c]{$2F$}}
\put(56,26){\makebox(4,3)[c]{$4N$}}
\put(56,23){\makebox(4,3)[c]{$5F$}}
\put(56,20){\makebox(4,3)[c]{$6F$}}

\put(85,30){\framebox(15,15)[c]{}} \put(90,30){\line(0,1){15}}
\put(95,30){\line(0,1){15}} \put(86,36){\makebox(4,3)[c]{$3F$}}
\put(91,40){\makebox(4,3)[c]{$2F$}}
\put(91,36){\makebox(4,3)[c]{$3F$}}
\put(91,33){\makebox(4,3)[c]{$4N$}}
\put(91,30){\makebox(4,3)[c]{$5F$}}
\put(96,36){\makebox(4,3)[c]{$3F$}}

\put(85,11){\framebox(15,12)[c]{}} \put(90,11){\line(0,1){12}}
\put(95,11){\line(0,1){12}} \put(86,16){\makebox(4,3)[c]{$6F$}}
\put(96,11){\makebox(4,3)[c]{$6F$}}
\put(96,14){\makebox(4,3)[c]{$5F$}}
\put(96,17){\makebox(4,3)[c]{$4N$}}
\put(96,20){\makebox(4,3)[c]{$2F$}}
\put(91,16){\makebox(4,3)[c]{$6F$}}

\put(0,27){\vector(1,0){4}} \put(20,27){\vector(1,0){25}}
\put(20,27){\vector(2,1){25.3}} \put(20,27){\vector(2,-1){25.5}}
\put(60,12){\vector(1,0){25}} \put(60,21){\vector(1,0){24}}
\put(60,42){\vector(1,0){25}} \put(60,33){\vector(1,0){25}}

\put(64,27){\circle{8}} \put(68,27){\vector(0,-1){1}}
\put(104,36){\circle{8}} \put(104,40){\vector(-1,0){1}}
\put(104,15){\circle{8}} \put(104,19){\vector(-1,0){1}}

\put(34,29){\makebox(0,0)[c]{$(a, a, a)$}}
\put(30,37){\makebox(0,0)[c]{$(c, \epsilon, c)$}}
\put(30,18){\makebox(0,0)[c]{$(b, b, \epsilon)$}}

\put(70,45){\makebox(0,0)[c]{$(a, a, a)$}}
\put(70,35){\makebox(0,0)[c]{$(c, \epsilon, c)$}}
\put(70,18){\makebox(0,0)[c]{$(b, b, \epsilon)$}}
\put(75,27){\makebox(0,0)[c]{$(a, a, a)$}}
\put(70,14){\makebox(0,0)[c]{$(a, a, a)$}}

\put(109,43){\makebox(0,0)[c]{$(a, a, a)$}}
\put(109,22){\makebox(0,0)[c]{$(a, a, a)$}}

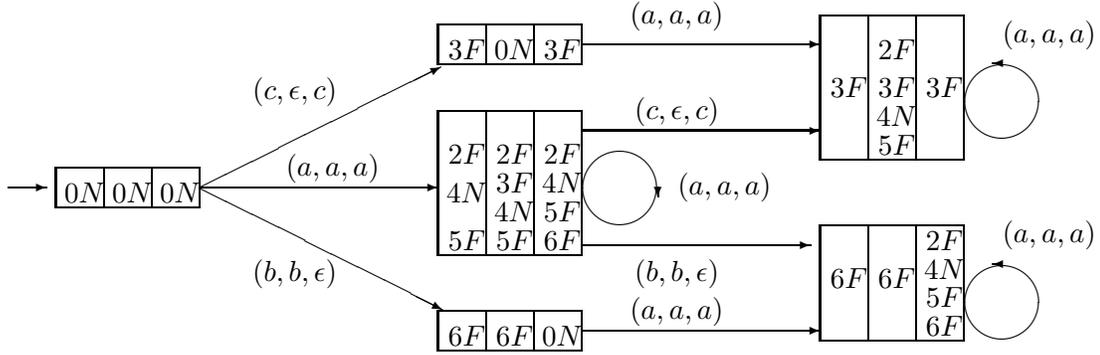
\put(40,2){\makebox(30,3)[c]{{\footnotesize Fig. 5. Codiagnoser
$G^{T}$ in Example 3.}}}

\end{picture}

From Fig. 5, we know that there are three cycles $C^{T}_{1},
C^{T}_{2}, C^{T}_{3}$ in $G^{T}$ as follows:
\begin{equation}
C^{T}_{1}=\left\{q^{T}_{2}, (a, a, a), q^{T}_{2}\right\}, \hskip
5mm C^{T}_{2}=\left\{q^{T}_{3}, (a, a, a),
q^{T}_{3}\right\},\hskip 5mm C^{T}_{3}=\left\{q^{T}_{5}, (a, a,
a), q^{T}_{5}\right\},
\end{equation}
 where
$$q^{T}_{2}=(\left\{(3,F)\right\}, \left\{(2,F), (3,F), (4,N),
(5,F)\right\}, \left\{(3,F)\right\}),$$$$q^{T}_{3}=(\left\{(2,F),
(4,N), (5,F)\right\}, \left\{(2,F), (3,F), (4,N), (5,F)\right\},
\left\{(2,F), (4,N), (5,F), (6,F)\right\}),$$
$$q^{T}_{5}=(\left\{(6,F)\right\}, \left\{(6,F)\right\},
\left\{(2,F), (4,N), (5,F), (6,F)\right\}).$$

In cycle $C^{T}_{1}$, state $q^{T}_{2}$ contains a uniform
recurrent component $(\left\{(3,F)\right\}, (3,F), (3,F))$ bearing
the label $F$, but it is not $F-$uncertain in $G_{T}$. In cycle
$C^{T}_{2}$, state $q^{T}_{3}$ is an $F-$uncertain state of
$G_{T}$, but it does not contain any uniform recurrent component.
Likewise, in cycle $C^{T}_{3}$, state $q^{T}_{5}$ contains a
uniform recurrent component $(\left\{(6,F)\right\}, (6,F), (6,F))$
bearing the label $F$, but it is not $F-$uncertain in $G_{T}$.
Therefore, there does not exist the cycle whose each state with a
uniform recurrent component bearing the label $F$ is
$F-$uncertain. By Theorem 1, we know that $L$ is codiagnosable,
which coincides with the result of Example 1. \hfill $\Box$

{\it Example 4.} Consider the stochastic automaton
$G=(Q,\Sigma,\eta, q_{0})$ represented by Fig.6, where
$Q=\{q_{0},\ldots, q_{6}\}$, $\Sigma=\{a, b, c, d, \sigma_{uo},
\sigma_{f}\}$, and $\Sigma_{f}=\{\sigma_{f}\}$. $L=L(G)$. Assume
there are two projections $P_{i}:\Sigma^{*}\rightarrow
\Sigma_{o,i}^{*}$, (where $i=1, 2$), $\Sigma_{o, 1}=\{a, b\}$,
$\Sigma_{o, 2}=\{a, d\}$.

\setlength{\unitlength}{0.1cm}
\begin{picture}(150,62)

\put(5,40){\circle{5}\makebox(-10,0){$0$}}
\put(25,40){\circle{5}\makebox(-10,0){$1$}}
\put(27,15){\circle{5}\makebox(-10,0){$5$}}
\put(44,50){\circle{5}\makebox(-10,0){$2$}}
\put(44,31){\circle{5}\makebox(-10,0){$4$}}
\put(47,15){\circle{5}\makebox(-10,0){$6$}}
\put(65,50){\circle{5}\makebox(-10,0){$3$}}

\put(-2,40){\vector(1,0){4}} \put(8,40){\vector(1,0){14}}
\put(5,37.5){\vector(1,-1){20.5}}
\put(29.5,15){\vector(1,0){14.5}}
\put(27.5,41){\vector(2,1){14.5}} \put(27.5,39){\vector(2,-1){14}}
\put(46.5,50){\vector(1,0){16}}

\put(44,56.5){\circle{8}} \put(39.7,56){\vector(0,-1){1}}
\put(71.5,50){\circle{8}} \put(71.5,45.5){\vector(-1,0){1}}
\put(50.5,31){\circle{8}} \put(54.5,31){\vector(0,-1){1}}
\put(31,20.5){\circle{8}} \put(35,20.5){\vector(0,-1){1}}
\put(53.5,15){\circle{8}} \put(53.5,11){\vector(-1,0){1}}

\put(15,43){\makebox(0,0)[c]{$(d, 0.7)$}}
\put(42.5,43){\makebox(0,0)[c]{$(\sigma_{f}, 0.8)$}}
\put(30,32){\makebox(0,0)[c]{$(\sigma_{uo}, 0.2)$}}
\put(9,25){\makebox(0,0)[c]{$(\sigma_{f}, 0.3)$}}
\put(42,20.5){\makebox(0,0)[c]{$(a, 0.8)$}}
\put(36,12){\makebox(0,0)[c]{$(b, 0.2)$}}
\put(33,56){\makebox(0,0)[c]{$(a, 0.7)$}}
\put(72,43){\makebox(0,0)[c]{$(a, 1)$}}
\put(61,31){\makebox(0,0)[c]{$(a, 1)$}}
\put(54,8){\makebox(0,0)[c]{$(a, 1)$}}
\put(56,53){\makebox(0,0)[c]{$(c, 0.3)$}}

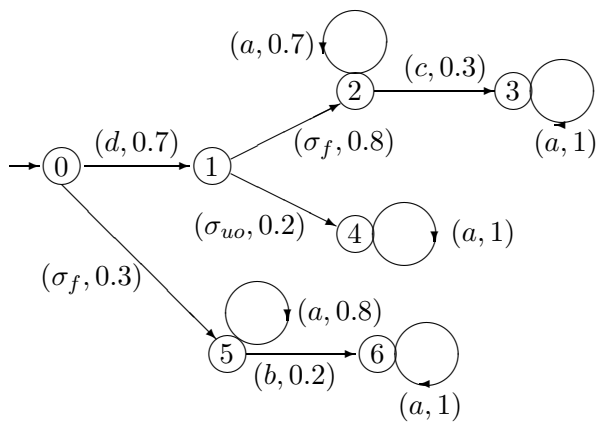
\put(25,1){\makebox(25,1)[c]{{\footnotesize Fig. 6. Stochastic
automaton of Example 4. }}}

\put(95,30){\framebox(6,6)[c]{$0N$}}
\put(110,45){\framebox(6,6)[c]{$5F$}}
\put(120,30){\framebox(6,6)[c]{$6F$}}
\put(110,15){\framebox(6,6)[c]{$1N$}}
\put(127,10){\framebox(6,13)[c]{}}
\put(127,10){\makebox(6,5)[c]{$4N$}}
\put(127,14){\makebox(6,5)[c]{$3F$}}
\put(127,18){\makebox(6,5)[c]{$2F$}}

\put(90,33){\vector(1,0){5}} \put(101,33){\vector(1,0){19}}
\put(100,36){\vector(1,1){10}} \put(116,45){\vector(1,-1){9}}
\put(100,30){\vector(1,-1){10}} \put(116,18){\vector(1,0){10.5}}

\put(113,55.3){\circle{8}} \put(117,55.3){\vector(0,-1){1}}
\put(130.3,33){\circle{8}} \put(134.3,33){\vector(0,-1){1}}
\put(137.3,15){\circle{8}} \put(137.3,19.1){\vector(-1,0){1}}

\put(102,41){\makebox(0,0)[c]{$a$}}
\put(111,35){\makebox(0,0)[c]{$b$}}
\put(102,23){\makebox(0,0)[c]{$d$}}
\put(124,41){\makebox(0,0)[c]{$b$}}
\put(119,55){\makebox(0,0)[c]{$a$}}
\put(136,33){\makebox(0,0)[c]{$a$}}
\put(120,20){\makebox(0,0)[c]{$a$}}
\put(137,21){\makebox(0,0)[c]{$a$}}

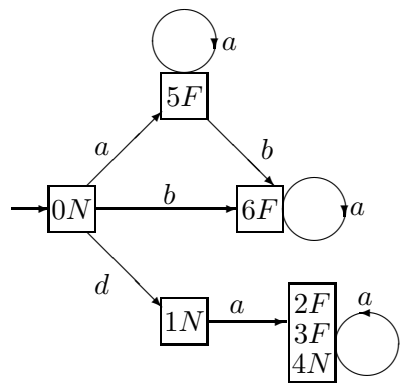
\put(110,1){\makebox(12,1)[c]{{\footnotesize Fig. 7. Diagnoser
$G^{'}_{D}$ in Example 4.}}}

\end{picture}

We assert that $L$ is not codiagnosable with respect to
$\left\{P_{i}: i=1,2\right\}$. In fact, we can verify this
conclusion by two scenarios as follows.

On the one side, we use the definition of codiagnosability of
SDESs (i.e., Definition 4) to interpret $L$ to be not
$F$-codiagnosable with respect to $\left\{P_{i}: i=1,2\right\}$.
We take $\epsilon=0.2$, $s=d\sigma_{f}\in \Psi(\Sigma_{f})$, and
$t=aca^{n-2}\in L/s $, then
$$ P_{1}^{-1}[P_{1}(st)]=\left\{d\sigma_{uo}a^{n-1},
\sigma_{f}a^{n-1}, d\sigma_{f}a^{n-1}, d\sigma_{f}a^{n-k-1}ca^{k}:
0 \leq k \leq n-1\right\},$$
$$ P_{2}^{-1}[P_{2}(st)]=\left\{d\sigma_{uo}a^{n-1},
d\sigma_{f}a^{n-1}, d\sigma_{f}a^{n-k-1}ca^{k}: 0 \leq k \leq n-1
\right\}.$$ Notice that $\sigma_{f}\notin d\sigma_{uo}a^{n-1}$, so
$D_{1}(st)=0$ and $D_{2}(st)=0$. However,
\begin{equation}
Pr(t:D_{1}(st)=0 \mid t\in L/s \wedge
\parallel t
\parallel=n)=Pr(aca^{n-2})=0.7\times 0.3=0.21>\epsilon,
\end{equation}
\begin{equation}
Pr(t:D_{2}(st)=0 \mid t\in L/s \wedge
\parallel t
\parallel=n)=Pr(aca^{n-2})=0.7\times 0.3=0.21>\epsilon.
\end{equation}
Therefore, $L$ is neither diagnosable with respect to $P_{1}$ nor
diagnosable with respect to $P_{2}$. By Definition 4 or
Proposition 2, we obtain that $L$ is not codiagnosable.

On the other side, we also can use Theorem 1 to verify that $L$ is
not codiagnosable.

According to the global projection $P:\Sigma^{*}\rightarrow
\Sigma_{o}^{*}$ and the local projections
$P_{i}:\Sigma^{*}\rightarrow \Sigma_{o,i}^{*}$, ($i=1,2$), we can
construct the diagnoser $G^{'}_{D}=(Q_{D},\Sigma_{o}, \delta_{D},
\chi_{0})$ for DFA $G^{'}$ as Fig. 7, and two local stochastic
diagnosers $G_{d}^{1}, G_{d}^{2}$ as Fig. 8, where
$\Sigma_{o}=\Sigma_{o,1}\cup\Sigma_{o,2}=\{a, b, d\}$, and
$$ G_{d}^{1}=(Q_{1}, \Sigma_{o,1}, \delta_{1}, \chi_{0}, \Phi_{1},
\phi_{0}), \hskip 6mm G_{d}^{2}=(Q_{2}, \Sigma_{o,2}, \delta_{2},
\chi_{0}, \Phi_{2}, \phi_{0}).
$$

\setlength{\unitlength}{0.05in}

\begin{picture}(130,41)
\put(5,30){\framebox(5,4)[c]{$0N$}}
\put(13,10){\framebox(5,4)[c]{$6F$}}
\put(35,25){\framebox(5,12)[c]{}}

\put(35,31){\makebox(5,3)[c]{$3F$}}
\put(35,34){\makebox(5,3)[c]{$2F$}}
\put(35,28){\makebox(5,3)[c]{$4N$}}
\put(35,25){\makebox(5,3)[c]{$5F$}}

\put(0,32){\vector(1,0){5}} \put(10,32){\vector(1,0){24.8}}
\put(7,30){\vector(1,-2){8}} \put(36,24.7){\vector(-2,-1){20.3}}

\put(16,7){\circle{6}} \put(18.8,7.5){\vector(0,-1){1}}
\put(37,39.7){\circle{6}} \put(34.2,39.7){\vector(0,-1){1}}

\put(2,34){\makebox(0,0)[c]{$\phi_{0}$}}
\put(22,34){\makebox(0,0)[c]{$(a, \phi^{1}_{1})$}}
\put(29,40){\makebox(0,0)[c]{$(a, \phi^{1}_{4})$}}
\put(7,22){\makebox(0,0)[c]{$(b, \phi^{1}_{2})$}}
\put(29,18){\makebox(0,0)[c]{$(b, \phi^{1}_{5})$}}
\put(24,7){\makebox(0,0)[c]{$(a, \phi^{1}_{3})$}}

\put(40,1){\makebox(30,1)[c]{{\footnotesize Fig. 8. Local
diagnosers $G^{1}_{d}$ (left one) and $G^{2}_{d}$ (right one) in
Example 4.}}}

\put(60,30){\framebox(5,4)[c]{$0N$}}
\put(70,10){\framebox(5,4)[c]{$1N$}}
\put(80,30){\framebox(5,7)[c]{}}
\put(80,33){\makebox(5,3)[c]{$5F$}}
\put(80,30){\makebox(5,3)[c]{$6F$}}
\put(90,7){\framebox(5,10)[c]{}}
\put(90,7){\makebox(5,3)[c]{$4N$}}
\put(90,10){\makebox(5,3)[c]{$3F$}}
\put(90,13){\makebox(5,3)[c]{$2F$}}

\put(55,32){\vector(1,0){5}} \put(65,32){\vector(1,0){14.8}}
\put(62,30){\vector(1,-2){8}} \put(75,12){\vector(1,0){15}}

\put(88,34){\circle{6}} \put(90.5,34){\vector(0,-1){1}}
\put(98,12){\circle{6}} \put(98,9.2){\vector(-1,0){1}}

\put(57,34){\makebox(0,0)[c]{$\phi_{0}$}}
\put(73,34){\makebox(0,0)[c]{$(a, \phi^{2}_{1})$}}
\put(96,34){\makebox(0,0)[c]{$(a, \phi^{2}_{4})$}}
\put(71,23){\makebox(0,0)[c]{$(d, \phi^{2}_{2})$}}
\put(82,15){\makebox(0,0)[c]{$(a, \phi^{2}_{3})$}}
\put(100,7){\makebox(0,0)[c]{$(a, \phi^{2}_{5})$}}

\end{picture}

For the local stochastic diagnoser $G_{d}^{1}$, the set of
probability transition matrices $\Phi_{1}=\left\{\phi_{0},
\phi^{1}_{1}, \phi^{1}_{2}, \phi^{1}_{3}, \phi^{1}_{4},
\phi^{1}_{5} \right\}$, where $\phi_{0}=[1], \hskip 3mm
\phi^{1}_{1}=[0.392, 0.168, 0.14, 0.24], \hskip 3mm
\phi^{1}_{2}=[0.06], \hskip 3mm \phi^{1}_{3}=[1]$,
$$\phi^{1}_{4}= \left[
\begin{array}{cccc}
0.7 & 0.3& 0 & 0\\
0& 1 & 0& 0\\
0& 0& 1& 0\\
0 & 0& 0 & 0.8
\end{array}
\right], \hskip 3mm \phi^{1}_{5}=\left[
\begin{array}{c}
0\\
0\\
0\\
0.2
\end{array}
\right].
$$
Therefore, the recurrent component bearing $F$ are $(q^{1}_{2}, 6,
F)$ and $(q^{1}_{3}, 3, F)$, where
$q^{1}_{2}=\left\{(6,F)\right\}$ and $q^{1}_{3}=\left\{(2,F),
(3,F), (4,N), (5,F)\right\})$. Notice that $q^{1}_{3}$ is not an
$F-$certain state of $G_{d}^{1}$, so $L$ is not diagnosable with
respect to $P_{1}$ by Lemma 3.

For the local stochastic diagnoser $G_{d}^{2}$, the set of
probability transition matrices $\Phi_{2}=\left\{\phi_{0},
\phi^{2}_{1}, \phi^{2}_{2}, \phi^{2}_{3}, \phi^{2}_{4},
\phi^{2}_{5} \right\}$ where $\phi_{0}=[1], \hskip 3mm
\phi^{2}_{1}=[0.24, 0.06], \hskip 3mm \phi^{2}_{2}=[0.7], \hskip
3mm \phi^{2}_{3}=[0.56, 0.24, 0.2]$,
$$\phi^{2}_{4}= \left[
\begin{array}{cc}
0.8 & 0.2\\
0& 1
\end{array}
\right], \hskip 3mm \phi^{2}_{5}=\left[
\begin{array}{ccc}
0.7 & 0.3 & 0\\
0 & 1 & 0\\
0 & 0 & 1
\end{array}
\right].
$$
Therefore,  the recurrent component bearing $F$ are $(q^{2}_{2},
6, F)$ and $(q^{2}_{4}, 3, F)$, where $q^{2}_{2}=\left\{(5,F),
(6,F)\right\}$ and $q^{2}_{4}=\left\{(2,F), (3,F),
(4,N)\right\})$. Notice that $q^{2}_{4}$ is not $F-$certain in
$G_{d}^{2}$, so $L$ is not diagnosable with respect to $P_{2}$ by
Lemma 3, either.

\setlength{\unitlength}{0.05in}

\begin{picture}(150,43)
\put(5,22){\framebox(15,4)[c]{}} \put(10,22){\line(0,1){4}}
\put(15,22){\line(0,1){4}} \put(6,22){\makebox(4,3)[c]{$0N$}}
\put(11,22){\makebox(4,3)[c]{$0N$}}
\put(16,22){\makebox(4,3)[c]{$0N$}}

\put(45,37){\framebox(15,4)[c]{}} \put(50,37){\line(0,1){4}}
\put(55,37){\line(0,1){4}} \put(46,37){\makebox(4,3)[c]{$1N$}}
\put(51,37){\makebox(4,3)[c]{$0N$}}
\put(56,37){\makebox(4,3)[c]{$1N$}}

\put(45,7){\framebox(15,4)[c]{}} \put(50,7){\line(0,1){4}}
\put(55,7){\line(0,1){4}} \put(46,7){\makebox(4,3)[c]{$6F$}}
\put(51,7){\makebox(4,3)[c]{$6F$}}
\put(56,7){\makebox(4,3)[c]{$0N$}}

\put(55,17){\framebox(15,12)[c]{}} \put(60,17){\line(0,1){12}}
\put(65,17){\line(0,1){12}} \put(56,22){\makebox(4,3)[c]{$5F$}}
\put(61,26){\makebox(4,3)[c]{$2F$}}
\put(61,23){\makebox(4,3)[c]{$3F$}}
\put(61,20){\makebox(4,3)[c]{$4N$}}
\put(61,17){\makebox(4,3)[c]{$5F$}}
\put(66,25){\makebox(4,3)[c]{$5F$}}
\put(66,20){\makebox(4,3)[c]{$6F$}}

\put(85,31){\framebox(15,12)[c]{}} \put(90,31){\line(0,1){12}}
\put(95,31){\line(0,1){12}} \put(86,31){\makebox(4,3)[c]{$4N$}}
\put(86,35){\makebox(4,3)[c]{$3F$}}
\put(86,39){\makebox(4,3)[c]{$2F$}}
\put(91,40){\makebox(4,3)[c]{$2F$}}
\put(91,37){\makebox(4,3)[c]{$3F$}}
\put(91,34){\makebox(4,3)[c]{$4N$}}
\put(91,31){\makebox(4,3)[c]{$5F$}}
\put(96,31){\makebox(4,3)[c]{$4N$}}
\put(96,35){\makebox(4,3)[c]{$3F$}}
\put(96,39){\makebox(4,3)[c]{$2F$}}

\put(85,8){\framebox(15,12)[c]{}} \put(90,8){\line(0,1){12}}
\put(95,8){\line(0,1){12}} \put(86,13){\makebox(4,3)[c]{$6F$}}
\put(96,9){\makebox(4,3)[c]{$6F$}}
\put(96,14){\makebox(4,3)[c]{$5F$}}
\put(91,13){\makebox(4,3)[c]{$6F$}}

\put(0,24){\vector(1,0){4}} \put(20,24){\vector(1,0){35}}
\put(20,24){\vector(2,1){25}} \put(20,24){\vector(2,-1){25.5}}
\put(60,9){\vector(1,0){25}} \put(60,38){\vector(1,0){24}}
\put(70,18){\vector(1,0){15}}

\put(74,24){\circle{8}} \put(78,24){\vector(0,-1){1}}
\put(104,37){\circle{8}} \put(104,33){\vector(-1,0){1}}
\put(104,12){\circle{8}} \put(104,16){\vector(-1,0){1}}

\put(38,26){\makebox(0,0)[c]{$(a, a, a)$}}
\put(30,34){\makebox(0,0)[c]{$(d, \epsilon, d)$}}
\put(30,15){\makebox(0,0)[c]{$(b, b, \epsilon)$}}

\put(70,40){\makebox(0,0)[c]{$(a, a, a)$}}
\put(78,15){\makebox(0,0)[c]{$(b, b, \epsilon)$}}
\put(85,24){\makebox(0,0)[c]{$(a, a, a)$}}
\put(70,11){\makebox(0,0)[c]{$(a, a, a)$}}

\put(109,30){\makebox(0,0)[c]{$(a, a, a)$}}
\put(109,19){\makebox(0,0)[c]{$(a, a, a)$}}

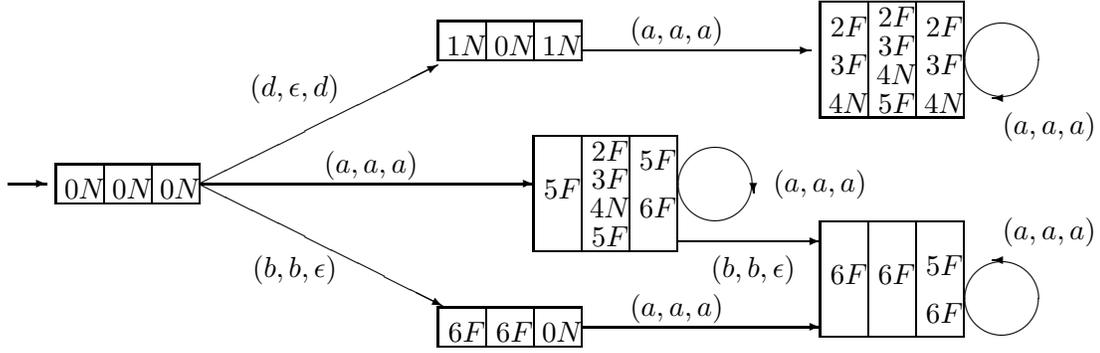
\put(40,1){\makebox(30,1)[c]{{\footnotesize Fig. 9. Codiagnoser
$G^{T}$ in Example 4.}}}

\end{picture}

Now we construct the codiagnoser $G_{T}$ to verify that $L$ is
also not codiagnosable.  The codiagnoser is a DFA $G_{T}=(Q_{T},
\Sigma_{T}, \delta_{T}, q^{T}_{0})$ as Fig. 9, where
$\Sigma_{T}=\left\{(a, a, a), (b, b, \epsilon), (d, \epsilon,
d)\right\}$.

From Fig. 9, we know that there are three cycles $C^{T}_{1},
C^{T}_{2}, C^{T}_{3}$ in $G^{T}$ as follows:
\begin{equation}
C^{T}_{1}=\left\{q^{T}_{2}, (a, a, a), q^{T}_{2}\right\},\hskip
5mm C^{T}_{2}=\left\{q^{T}_{3}, (a, a, a),
q^{T}_{3}\right\},\hskip 5mm C^{T}_{3}=\left\{q^{T}_{5}, (a, a,
a), q^{T}_{5}\right\},
\end{equation}
where $$q^{T}_{2}=(\left\{(2,F), (3,F), (4,N)\right\},
\left\{(2,F), (3,F), (4,N), (5,F)\right\}, \left\{(2,F), (3,F),
(4,N)\right\}),$$
$$q^{T}_{3}=(\left\{(5,F)\right\}, \left\{(2,F), (3,F),
(4,N), (5,F)\right\}, \left\{(5,F), (6,F)\right\}),$$
$$ q^{T}_{5}=(\left\{(6,F)\right\}, \left\{(6,F)\right\},
\left\{(5,F), (6,F)\right\}).$$

In the cycle $C^{T}_{1}$, there is only one state $q^{T}_{2}$ and
it contains a uniform recurrent component
$$(\left\{(2,F), (3,F), (4,N)\right\}, (3,F), (3,F))$$ bearing the
label $F$. Furthermore,  $q^{T}_{2}$ is an $F-$uncertain state of
$G_{T}$. Therefore, there does exist a cycle (i.e., $C^{T}_{1}$)
whose each state with a uniform recurrent component bearing the
label $F$ is $F-$uncertain. By Theorem 1, we obtain that $L$ is
not codiagnosable. \hfill $\Box$

\section*{V.  Concluding Remarks}
Recently,  J. Lunze and J. Schr\"{o}der [16], D. Thorsley and D.
Teneketzis [31] generalized the diagnosability of classical DESs
[25, 26] to the setting of stochastic DESs (SDESs). In [16], the
diagnostic problem was transformed into an observation problem,
and the diagnosability was obtained by an extension of an
observation algorithm. In [31], the notions of A- and
AA-diagnosability for stochastic automata were defined, which were
weaker than those for classical automata introduced by Sampath
{\it et al} [25, 26], and they [31] presented a necessary and
sufficient condition for the diagnosability of SDESs.

However, the failure diagnosis they considered in [16, 31] was
still centralized. In this paper, we have dealt with the
decentralized failure diagnosis for SDESs. The centralized failure
diagnosis of SDESs in [16, 31] can be viewed as a special case of
the decentralized failure diagnosis presented in this paper with
only one projection. We formalized the approach to decentralized
failure diagnosis by introducing the notion of codiagnosability.
By constructing a codiagnoser from a given stochastic automaton
with multiple projections, we used the codiagnoser associated with
the local diagnosers to test codiagnosability condition of SDESs.
As well, a number of basic properties of the codiagnoser has been
investigated. In particular, a necessary and sufficient condition
for the codiagnosability of SDESs was presented, which generalizes
the result of classical DESs dealt with by W. Qiu and R. Kumar
[21]. Furthermore, we gave a computing method in detail to check
whether codiagnosability is violated. Finally, some examples were
described to illustrate the applications of the codiagnosability
and its computing method.

The problem of decentralized diagnosis can be considered as one
special case of distributed diagnosis in [7]. Therefore, the
potential of applications of the results in this paper may be in
 failure diagnosis of many large complex systems which are
physically distributed [7, 21, 22, 28].  Moreover, with the
results obtained in this paper, a further issue worthy of
consideration is the strong codiagnosability of SDESs, as the
strong codiagnosability of classical DESs [21]. Another important
issue is how to compute the bound in the delay of decentralized
diagnosis for SDESs. We would like to consider them in subsequent
work.

\end{document}